\documentclass{ieeetj}
\usepackage{cite}
\usepackage{amsmath,amssymb,amsfonts}
\usepackage{algorithmic}
\usepackage{graphicx,color}
\usepackage{textcomp}
\usepackage{xcolor}
\usepackage{tabularx} %
\usepackage{hyperref}
\usepackage{dirtytalk}
\usepackage{booktabs}
\usepackage{multirow}
\usepackage{lipsum}
\hypersetup{hidelinks=true}
\usepackage{algorithm,algorithmic}
\def\BibTeX{{\rm B\kern-.05em{\sc i\kern-.025em b}\kern-.08em
    T\kern-.1667em\lower.7ex\hbox{E}\kern-.125emX}}
\AtBeginDocument{\definecolor{tmlcncolor}{cmyk}{0.93,0.59,0.15,0.02}\definecolor{NavyBlue}{RGB}{0,86,125}}

\def\OJlogo{\vspace{-4pt}$<$Society logo(s) and publication title will appear here.$>$}
\def\seclogo{\vspace{10pt}$<$Society logo(s) and publication title will appear here.$>$}

\def\authorrefmark#1{\ensuremath{^{\textbf{#1}}}}

\begin{document}
\receiveddate{XX Month, XXXX}
\reviseddate{XX Month, XXXX}
\accepteddate{XX Month, XXXX}
\publisheddate{XX Month, XXXX}
\currentdate{XX Month, XXXX}
\doiinfo{XXXX.2022.1234567}

\markboth{The Coastline as a Structural Constraint: Harnessing Scene Geometry for Autonomous Surface Vessel Localization}{Benham and Mangelson}

\title{The Coastline as a Structural Constraint: Harnessing Scene Geometry for Autonomous Surface Vessel Localization}

\author{Derek R. Benham\authorrefmark{1}, and Joshua G. Mangelson\authorrefmark{1} (Member, IEEE)}
\affil{Brigham Young University, Provo, UT 84602 USA}
\corresp{Corresponding author: Derek R. Benham (email: laser14@byu.edu).}
\authornote{This work was partially funded under Office of Naval Research award numbers N00014-24-1-2301 and N00014-24-1-2503.}

\begin{abstract}
Coastal environments contain rich, largely unexploited geometric structure capable of providing globally referenced localization cues. In this work, we present two complementary localization frameworks that exploit shoreline and water-surface geometry for GPS-denied autonomous surface vessel localization.
The first framework leverages LiDAR observations of the water surface to estimate roll, pitch, and heave (vertical motion), while recovering global position and heading through direct registration of shoreline observations against a satellite-derived coastline map.
The second framework relies solely on passive imagery to detect the shoreline and horizon through semantic segmentation. 
Using the proposed coastal scene geometry, shoreline distance is inferred from monocular imagery. Shoreline observations are accumulated into short-duration local submaps, registered against the same satellite-derived coastline map, and fused within a hierarchical factor graph.
Evaluated across three real-world coastal datasets, the LiDAR pipeline consistently improves trajectory accuracy over standard baselines, while the monocular architecture maintains bounded long-term drift. 
In addition, we establish that modern zero-shot foundation models can reliably extract shoreline observations across diverse coastal environments.
Together, these results demonstrate that coastal geometry provides a powerful and dependable source of globally referenced information for GPS-denied maritime localization.
\end{abstract}


\maketitle

\section{INTRODUCTION}
\label{sec:intro}

\IEEEPARstart{T}{he} demand for Autonomous Surface Vessels (ASVs) is increasing as these versatile, unmanned systems enable extended operations throughout the marine domain. Their applications are broad, ranging from maritime infrastructure inspection and coastline management to coastal water security and deterrence. A robust localization system is fundamental for these systems to support intelligent behaviors, such as patrolling specific regions or generating high-fidelity maps of marine environments or structures.

Despite their expanding capabilities, many current systems remain heavily reliant on GPS. While some platforms incorporate dead-reckoning filters to provide resilience during short-term outages, they ultimately drift and fail without regular global position estimates. 
One alternative to recover global position without GPS is to localize into a pre-existing map of the environment, but this restricts operation to mapped regions and ties localization accuracy to the fidelity and recency of the reference map.

Simultaneous Localization and Mapping (SLAM) was developed to eliminate the dependence on prior maps by jointly estimating an agent's pose while simultaneously constructing a map of its environment. Yet without an external global reference, traditional SLAM systems can only generate maps with respect to a local origin. Aligning this local frame to a global coordinate system is non-trivial without prior knowledge of the environment or intermittent GPS measurements. Furthermore, as mission durations extend, the underlying graph structures in SLAM can become computationally prohibitive to optimize.

\begin{figure}
    \centering
    \includegraphics[width=0.95\columnwidth]{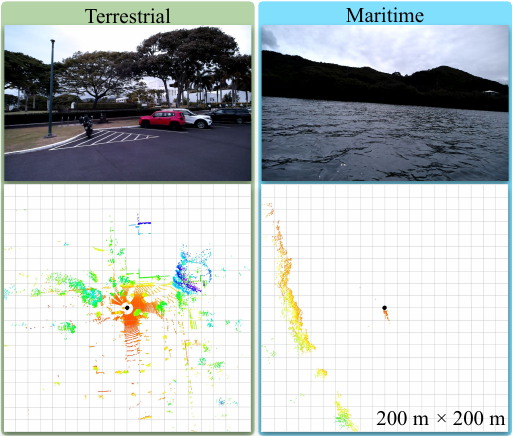}
    \caption{Comparison of terrestrial and maritime sensing environments, demonstrating the lack of traditional information present on water compared to land. Terrestrial environments naturally contain abundant nearby structure distributed throughout the sensor field of view, producing dense observations well suited for conventional localization methods. In contrast, the maritime environment is dominated by open water, with sparse observations of land occurring at greater distances and often on only one side of the vehicle. This disparity highlights the fundamental sensing challenges of maritime localization and motivates the need for domain-specific localization strategies.}
    \label{fig:coastal_challenges}
\end{figure}

One category of prior map that is publicly accessible, memory-efficient, globally referenced, and frequently updated is satellite imagery. Consequently, cross-view localization has emerged as an active research field focused on localizing ground-level observations within an overhead satellite view of the operational area~\cite{crossview_survey}. The effectiveness of such approaches ultimately depends on identifying environmental features that are both observable from the vehicle and persistent in overhead imagery. 
While significant progress has been made for autonomous driving in urban environments~\cite{shi2022beyond, wang2023satellite}, the maritime domain, and coastal localization in particular, remains largely under-explored~\cite{dagdilelis2025multimodalmultiviewdeepfusion}. As illustrated in Fig.~\ref{fig:coastal_challenges}, maritime environments present sensing conditions that differ markedly from those encountered on land. Nevertheless, many existing methods seek solutions that generalize across both terrestrial and maritime domains~\cite{rascl}, often overlooking environmental cues unique to coastal scenes.

Unlike most terrestrial environments, coastal scenes contain several naturally occurring geometric features that remain stable over time, are observable across sensing modalities, and can provide globally referenced information.
The mean water surface defines a globally consistent reference plane, the horizon provides absolute attitude information, and the shoreline forms a persistent environmental boundary observable from both ground-level sensors and overhead satellite imagery. We will show that, together, these complementary structures enable globally referenced localization without requiring artificial infrastructure, learned landmarks, or previously reconstructed maps.

Building upon these structural properties of the coastline, we develop two complementary localization pipelines that exploit coastal scene geometry through LiDAR and monocular vision. Our LiDAR pipeline combines water-surface plane estimation with cross-view shoreline registration to formulate globally referenced six-degree-of-freedom pose observations. 
Our passive monocular vision pipeline uses semantically labeled images to extract both horizon and shoreline observations. Horizon observations constrain roll and pitch, while shoreline observations are accumulated into short-duration submaps before registration against satellite imagery.

The vision pipeline presents two additional challenges. First, the coastline and horizon must be robustly extracted from monocular imagery despite changing illumination and diverse coastal landscapes. To address this, we employ a zero-shot foundation-model perception pipeline that performs semantic segmentation without requiring task-specific training data. Second, individual shoreline observations remain highly uncertain and cannot be registered reliably in isolation. Instead, observations are accumulated over extended time intervals to construct shoreline submaps with sufficient geometric information for robust satellite registration. Maintaining the consistency of these submaps requires accurate local trajectory estimation while preserving the ability to retroactively distribute delayed localization corrections. To address this challenge, we implement a hierarchical factor graph architecture that couples a high-rate fixed-lag local estimator with a sparse global pose graph, enabling computationally efficient localization over long-duration missions.

The primary contributions of this work include:
\begin{itemize}
    \item A unified coastal localization framework demonstrating how shoreline geometry can be exploited for ASV localization using either LiDAR or monocular vision.
    \item A LiDAR-based localization pipeline that combines water-surface plane estimation with cross-view shoreline alignment to recover globally referenced six-degree-of-freedom pose observations.
    \item A monocular vision pipeline that reconstructs shoreline submaps and performs observability-aware satellite registration within a hierarchical factor graph architecture for computationally efficient long-duration localization.
    \item An extensive experimental evaluation across three real-world coastal datasets totaling nearly 5~km of operation.
\end{itemize}

The remainder of this paper is organized as follows. Section \ref{sec:related_work} reviews prior work in LiDAR- and vision-based ASV localization, together with recent advances in coastal scene segmentation. Section \ref{sec:method} introduces the proposed coastal scene model and presents the LiDAR and monocular localization pipelines that exploit this geometry for state estimation. Section \ref{sec:results} presents experimental results for both the segmentation front-end and the proposed localization pipelines. Section \ref{sec:lessons_learned} discusses lessons learned and limitations identified through the experimental evaluation. Section \ref{sec:conclusion} summarizes the key contributions and concludes the paper.

\section{RELATED WORK} 
\label{sec:related_work}
Early research in ASV state estimation was largely contingent on the availability of GPS~\cite{wamv_waypoint}, often focusing on inland or sheltered waters where wave-induced roll, pitch, and heave were negligible~\cite{squirtle_2014, gps_waypoint}. To address more dynamic environments, Hitz et al. introduced decoupled estimators tracking pose in $SE(2)$ while estimating attitude with a complementary filter~\cite{hitz_hal-01174626}. Subsequent works advanced toward full six-degree-of-freedom ($6$-DOF) tracking using filtering techniques such as the Unscented Kalman Filter (UKF)~\cite{KONRAD2018181, low_cost_mems_usv} and the Invariant Extended Kalman Filter (InEKF)~\cite{benham2026invariantextendedkalmanfilter}.While these methods achieve high precision in attitude and short-term displacement tracking, they remain fundamentally reliant on GPS for global position anchoring. 
To address this limitation, the remainder of this section surveys GPS-denied localization approaches for ASVs, global cross-view techniques, and recent advances in semantic coastal perception.

\subsection{Coastal State Estimation and Odometry}
Unlike vessels operating in the open ocean, vessels navigating near shore can exploit observations of the surrounding coastline for localization.
Among modern autonomous vessels, radar, LiDAR, and cameras are the most commonly deployed above-water exteroceptive sensors.

\subsubsection{Radar}
Marine radar offers a sensing modality that is inherently robust to the adverse perceptual conditions of the ocean. Electromagnetic radar pulses are highly resilient to water-surface clutter and atmospheric obscurants such as fog and heavy spray~\cite{s21165397}. However, this robustness is typically accompanied by lower angular resolution and multipath interference, producing a coarser geometric representation of the environment than LiDAR. Despite these characteristics, radar-based odometry techniques have grown in prominence within the robotics community~\cite{burnett2022we, 9197231, 10611194, 10610311, 8764393}, particularly in maritime applications~\cite{moana} where radar has been ubiquitous for navigation and collision avoidance since the 1940’s~\cite{radar_history}.

To leverage the strengths of this modality, several recent works have focused on robust radar odometry and localization for ASVs. Monaco and Brennan~\cite{marine_radar_odom} proposed a two-phased motion estimation pipeline that matches raw radar signal returns to determine changes in heading, followed by scan-to-scan feature matching to estimate translational velocity. Similarly, Han et al.~\cite{marine_radar_slam} 
correlates radar returns with prior radar maps to bound dead-reckoning drift. 
To address scenarios where pre-existing maps are completely unavailable, \cite{8600301} developed a coastal SLAM algorithm that models the shoreline online using continuous splines generated from raw radar scans.

\subsubsection{LiDAR}
Compared with radar, LiDAR provides substantially denser and more precise observations of the surrounding environment, making it highly attractive for obstacle avoidance, scan registration, and coastal mapping. However, this high precision comes at the cost of reduced range and increased sensitivity to water-surface reflections and backscatter~\cite{lidar_obstacle}. Addressing these sensing challenges has led researchers to explore both specialized filtering techniques and LiDAR-based maritime navigation frameworks.

Early maritime localization efforts primarily used LiDAR to complement or bridge gaps in GNSS availability. In \cite{lidar_2d_marine}, a fusion positioning method was proposed that dynamically switches between GNSS/INS loosely coupled integration in open waters and a LiDAR-SLAM-assisted mode during GPS outages. Hitz et al.~\cite{Hitz2016} employed Iterative Closest Point (ICP) alignment~\cite{icp_seminal} to refine local position estimates and enhance vessel attitude tracking. 

A seminal LiDAR-inertial odometry work, LIO-SAM~\cite{liosam}, included evaluations of their method onboard a boat cruising through the canals of Amsterdam. This tightly coupled factor-graph-based framework achieved a translation error of only 0.17~m in environments where traditional methods such as LOAM~\cite{loam} and LIOM~\cite{liom} failed to produce meaningful results. The experimental environment was notably feature-rich, with buildings and canal walls on both sides of the vessel providing abundant structure for scan registration and localization, while the water remained relatively calm.

While still remaining on inland waters, Wang et al.~\cite{inland_gicp} proposed a LiDAR-based SLAM framework tailored for waterways that highlights the necessity for specialized point cloud preprocessing when operating in aquatic environments. Their approach utilizes a rejection sampling-based GICP to improve odometry robustness in feature-poor regions while incorporating a specialized preprocessing filter to mitigate water-surface noise. 

Despite improving local stability and data consistency, these approaches remain fundamentally dependent on GNSS measurements to establish global position, rather than leveraging environmental structure as a source of globally referenced localization.

\subsubsection{Cameras}
Due to their low cost and widespread availability, camera-based vision systems have become increasingly common on ASVs. Most applications focus on obstacle avoidance \cite{10314528, 7073635} and target tracking \cite{PARK2024100608, 9144189}, while comparatively less attention has been devoted to GPS-denied localization.

While some studies employ visual odometry (VO) or visual-inertial odometry (VIO) for state estimation, these applications are largely restricted to inland waterways~\cite{9381638}. Their performance is often highly dependent on shoreline proximity. For example, \cite{terzakis2017monocularvisualodometryunmanned} observed localization errors increase from approximately 1~m to over 20~m as the vessel moved from near-shore operation toward the center of a calm river, highlighting the sensitivity of traditional visual localization methods to stand-off distance.
This severe dependency on shoreline proximity highlights the difficulty of directly applying standard robotic state estimation algorithms to the marine domain. To demonstrate this mismatch, Filip et al.~\cite{10773240} evaluated VINS-Mono~\cite{vins_mono} and VINS-Stereo~\cite{qin2019a}, frameworks originally designed for aerial or ground vehicles, on a custom inland waterway dataset. While both configurations successfully constrained vertical ($z$-axis) drift, they accumulated substantial horizontal drift (209~m and 55~m, respectively) over a 2.5~km trajectory. 

Sensing in wide-open coastal or marine environments represents the extreme limit of this feature-dependency problem. To evaluate odometry at larger distances from sheltered shorelines, Nguyen et al.~\cite{NGUYEN2022235} investigated the feasibility of monocular Direct Sparse Odometry (DSO) for vessel state estimation. The authors concluded that while DSO was insufficient for stable position tracking, it could reliably capture relative rotational motion. 
Similar in philosophy to our work, Meier et al.~\cite{curve_slam} demonstrated that exploiting environment-specific geometric constraints can substantially improve localization performance. In calm river environments, they leverage the physics of specular reflections by identifying symmetric point pairs across the water surface to distinguish true scene features from mirror reflections. These constraints are incorporated into a factor graph through Bézier curve landmarks to improve attitude estimation and downstream localization.

Ultimately, these vision-based techniques ranging from pure visual odometry to local SLAM frameworks remain bound by environment-specific trade-offs. Whether restricted to inland waterways to maintain proximity to shore-based landmarks, or deployed in open-water environments where feature tracking quickly fails, these methods localize the vessel relative to an arbitrary, locally initialized coordinate frame. 

\subsection{Cross-View Localization}
Recovering global position requires registering the local map against a geo-referenced prior. Cross-view localization addresses this problem by aligning ground-level sensor observations with overhead representations, such as satellite imagery.

For instance, Ma et al.~\cite{7907227} utilized geo-referenced satellite imagery as a prior to align radar scans. Although their template-matching technique significantly outperformed dead reckoning, it aligned each radar scan independently without optimizing for temporal consistency, limiting its ability to generate a smooth, continuous trajectory between consecutive observations.

Building upon these concepts, Blerim et al.~\cite{rascl} introduced a cross-view localization framework that jointly optimized local radar odometry while simultaneously aligning observations with overhead satellite imagery. Their approach trained an attention-based U-Net to synthesize pseudo-radar scans from satellite imagery, enabling global localization by aligning real-time radar observations with the predicted scans. Restricted to the $SE(2)$ manifold and evaluated on the same lake used to train the network, their method achieved a positional RMSE of 3.5~m over a 650~m trajectory.

Beyond localization via satellite maps, other researchers have explored deep learning based visual geolocation (VG) methods~\cite{brest_dl_terrain, usv_terrain_localization}. These approaches use Convolutional Neural Networks (CNNs) to match camera imagery with digital elevation models of the coastal terrain, achieving sub–two-meter position accuracy in calm-water experiments. Crucially, these works focused solely on horizontal position accuracy, neglecting vessel attitude estimation and failing to integrate measurements within a consistent recursive or graph-based state estimator.

Collectively, while these methods demonstrate the effectiveness of prior geo-referenced maps for bounding drift, they remain largely limited to specific sensing modalities and do not fully exploit the broader geometric and semantic structures inherent to coastal environments.

\subsection{Semantic Coastal Scene Understanding}
Semantic segmentation has opened up new possibilities for robotic platforms to leverage high-level semantic scene understanding to refine state estimation~\cite{sh-ch16-semantic}. While fully autonomous operations may require semantic understandings of diverse structures, including buoys, other vessels, docks, marine debris, and shoreline obstacles, our proposed method exploits only the unique geometric constraints of coastal environments. Consequently, it requires robust segmentation of just three broad semantic classes: water, sky, and land. This section reviews the various approaches in the literature used to segment these three classes, covering four core paradigms: traditional computer vision, fully supervised CNNs, transfer learning approaches, and zero-shot foundation models.

Historically, traditional computer vision approaches have been favored for real-time maritime applications due to their low latency and minimal computational footprints. For instance, to filter out erroneous visual feature points, Zhang et al.~\cite{jmse13040679} utilized Otsu's thresholding method~\cite{otsu} to generate a land mask, thereby filtering out features detected on highly reflective water or dynamic sky regions. Meanwhile, Benham et al.~\cite{benham2026invariantextendedkalmanfilter} employed a line segment detector to locate the horizon line, utilizing this geometric boundary to infer vessel roll and pitch. While these traditional methods provide valuable geometric cues for specific tasks, they lack a holistic semantic understanding of the scene; they cannot identify the broader context of what is being masked or what a detected boundary line represents, whether it is a horizon or a coastline. This lack of situational context underscores the need for deep learning-based semantic segmentation to inform state estimation modules.

CNNs address these contextual limitations by providing simple architectures capable of learning rich features directly from data. Zhan et al.~\cite{s19102216} utilized a classic U-Net architecture to semantically segment the water-plane for autonomous navigation, operating under the assumption that any non-water region constitutes a potential obstacle. A major bottleneck of these fully supervised, learned approaches is their heavy reliance on massive, high-quality labeled datasets to achieve generalization across varying lighting and sea conditions.

To mitigate the high cost of training models from scratch, transfer learning offers a viable pathway to leverage representations learned by large-scale models and adapt them to specific downstream maritime tasks. For example, the authors of \cite{marine_vessel_attitude} extended a pre-trained, lightweight MobileNetV2 backbone~\cite{sandler2018mobilenetv2} with a custom U-Net decoder to estimate class-wise pixel probabilities. Similarly, YOLO provides an efficient segmentation model (YOLO-Seg) that is highly amenable to transfer learning~\cite{yolov8_ultralytics}.

To address the resource constraints of onboard edge devices, Ter\v{s}ek et al.~\cite{ewasr} developed eWaSR, an optimized WaSR~\cite{wasr} variant using a lightweight ResNet-18 backbone in place of the computationally intensive ResNet-101. This modification balances real-time efficiency with high segmentation accuracy on marine platforms, making the architecture highly attractive for field deployment. Maintaining this accuracy across varying geographic locations often requires location specific fine-tuning.

To address this environmental variability, Huang et al.~\cite{10648784} argue that the most critical factor in maintaining a robust neural network is a large and diverse dataset. They leverage the efficient architecture and pre-trained weights of eWaSR, expanding its exposure by introducing an additional 4,000 images tailored to their specific operational environment. Yet, while such targeted transfer learning improves local performance, it underscores a broader limitation in marine perception: in order for semantic segmentation to scale effectively without requiring bespoke datasets for every new deployment, a more fundamentally generalized approach is necessary.

Vision foundation models offer a promising pathway toward zero-shot generalization, yet a distinct capability gap remains between models optimized for dense semantic segmentation and those trained for open-vocabulary recognition. Standard Vision Transformer (ViT) classifiers can categorize complex scenes but lack the dense, pixel-level segmentation boundaries required for robotic mapping. This discrepancy motivates a decoupled two-stage paradigm in which a semantic frontend assigns open-vocabulary labels to an image while a high-fidelity spatial backend performs pixel-wise boundary extraction.

Huang et al.~\cite{10648784} originally employed a transfer learning-based model because, at the time, deep vision foundation models were computationally prohibitive for real-time edge execution. To bypass the bottleneck of manual data labeling, the authors paired the Segment Anything Model (SAM)~\cite{kirillov2023segment} with human-provided prompts, using SAM to automate water segmentation. Since their work, advances in both open-vocabulary classification and zero-shot segmentation have rendered real-time execution on physical edge devices feasible without human intervention~\cite{awais2025foundation}.

While the original SAM relies on a computationally heavy ViT backbone that introduces prohibitive latency for closed-loop navigation, FastSAM~\cite{zhao2023fast} and MobileSAM~\cite{zhang2023faster} reformulate the task as an instance segmentation problem utilizing a CNN. This architectural shift drastically reduces inference times while preserving robust, zero-shot boundary localization. For consistency in dynamic environments, SAM~2~\cite{ravi2025sam} introduced a critical advancement through its unified model design and memory-attention mechanism. By tracking object masks across continuous video streams rather than processing frames in isolation, SAM~2 maintains mask consistency and mitigates the pervasive issue of frame-to-frame mask flickering.

To semantically understand these classless masks, several frontend paradigms exist to interact with this spatial backend, fundamentally categorized by whether they operate via prompting or semantic inference. A prompting frontend uses natural language input to identify a target category and subsequently prompts the backend segmentation model to extract a corresponding pixel-level mask. Conversely, an inference-based frontend receives the class-agnostic segments generated by the backend and infers their semantic identity, assigning semantic labels to the resulting regions without explicit prompting.

Grounding DINO~\cite{liu2024grounding} and YOLO-World~\cite{cheng2024yolo} are two prominent prompting-based frontend models. While Grounding DINO leverages a dense Transformer-based architecture to achieve deep vision-language alignment, YOLO-World relies on a highly optimized CNN backbone designed for real-time edge execution. Despite these differing architectures, both models operate by detecting user-specified categories (e.g., water, sky, or land) and providing the resulting spatial coordinates as prompts to the backend segmentation model. Models such as CLIP~\cite{radford2021learning} represent the semantic inference paradigm, establishing a shared latent space between text and imagery that enables class-agnostic segments to be assigned semantic labels from arbitrary natural language descriptions.

As research in ViTs and Vision-Language Models continue to accelerate, recent works like YOLO-Everything (YOLO-E)~\cite{yolo_e} and SAM-3~\cite{carion2025sam} fuse the two-stage pipeline into a single, unified architecture capable of providing pixel-wise classification directly from an open-set language-based prompt. However, these monolithic, end-to-end models introduce critical trade-offs that complicate their practical deployment. For SAM-3, the immense computational overhead of unifying dense segmentation with complex language modeling comes at the price of inference speed and sacrificed modular flexibility. Unlike decoupled paradigms, users cannot independently upgrade or tune portions of the model as newer, more optimized models emerge. YOLO-E manages to maintain faster than real-time performance, but treats each image independently and can struggle with temporal consistency when operating on sequences of frames. 

These recent advances make semantic segmentation a practical frontend for our proposed localization framework. Accordingly, our vision pipeline leverages Grounding DINO with SAM~2 as a zero-shot segmentation model to extract the water, sky, and land observations that underpin the geometric constraints used for localization.

\section{METHODS}
\label{sec:method}

To leverage the unique structural properties of coastal environments, we propose two complementary localization methods that exploit drift-free geometric constraints derived from the shoreline and water surface. We first establish the mathematical model of the coastal scene. We then present Coastal-KISS, a LiDAR localization framework built upon KISS-ICP~\cite{vizzo2023kiss} that combines water-surface plane estimation with cross-view shoreline registration to recover globally referenced pose measurements. Finally, we introduce a vision-based localization method that combines semantically segmented imagery with inertial sensing. Shoreline distance is inferred through monocular geometric reconstruction and incorporated within a hierarchical factor graph that addresses the limited observability of coastal environments.

\subsection{Coastal Scene Modeling and Assumptions}
\begin{figure}
    \centering
    \includegraphics[width=\columnwidth]{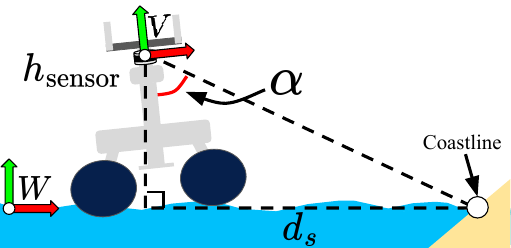}
    \caption{
    Geometric model of the coastal environment. Both the LiDAR and monocular vision pipelines assume a locally planar water surface with the coastline defined as the boundary between the water surface and land. The resulting right-triangle geometry between the vehicle, water surface, and shoreline provides simple geometric constraints for globally referenced localization, where $h_\text{sensor}$ denotes the sensor height above the water surface, $d_s$ the distance to shore, and $\alpha$ the observed declination angle to the shoreline.
    }
    \label{fig:coastline_assumptions}
\end{figure}

Both localization pipelines are derived from the same geometric model of the coastal environment, illustrated in Fig.~\ref{fig:coastline_assumptions}, despite relying on different sensing modalities. Let the global coordinate system be denoted by the world frame $W$, and the vehicle-affixed coordinate system be the sensor frame $V$. We model the mean water surface as a planar reference with elevation $z_W$, initialized from the tide level at mission start. Although the instantaneous water surface is perturbed by waves, its time-averaged elevation is assumed constant over the duration of a mission.

The coastline is modeled as the intersection of the terrain and the mean water surface. While the instantaneous shoreline position varies due to wave motion and wave run-up, these fluctuations are assumed to occur about a stable mean water level. Tidal variation acts on a much longer timescale and is therefore assumed negligible over the 15--20 minute duration of each mission. Under these assumptions, the mean water surface and coastline provide stationary geometric references that can be incorporated as localization constraints within both sensing frameworks.

\subsection{LiDAR based Localization}

\begin{figure*}
    \centering
    \includegraphics[width=\textwidth]{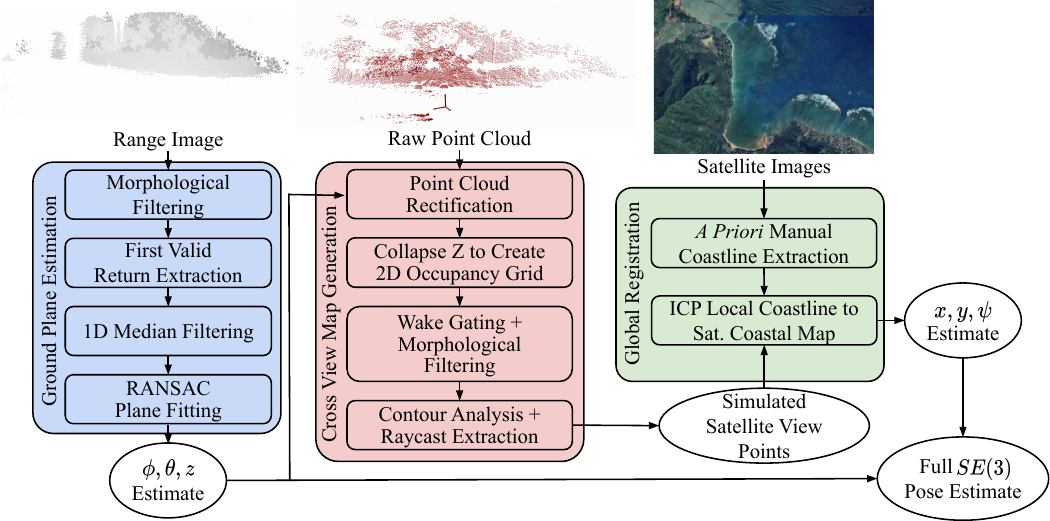}
    \caption{
    LiDAR localization overview. The pipeline consists of three main stages. For ground plane estimation, the native LiDAR range image is utilized to extract a local water-surface plane, providing absolute estimates of the vehicle's roll~($\phi$), pitch~($\theta$), and heave~($z$) with respect to the world frame. For cross-view map generation, the raw point cloud is rectified using these attitude estimates and orthographically projected to a 2D occupancy grid to extract a simulated satellite view of the coastline. Finally, for global registration, this simulated view is aligned via ICP to an \textit{a priori} map of the coastline to resolve the vehicle's planar ($x,y$) position and heading ($\psi$). Fusing these vertical and planar estimates yields a full $SE(3)$ unary observation factor for integration into a fixed-lag smoother factor graph.
    }
    \label{fig:lidar_pipeline}
\end{figure*}

Building upon the coastal scene model introduced in the previous section, the proposed LiDAR localization pipeline (Fig.~\ref{fig:lidar_pipeline}) exploits both the LiDAR range image and 3D point cloud to recover globally referenced pose measurements. The range image is used to estimate roll, pitch, and height from the observed water surface, while shoreline geometry extracted from the point cloud is registered against a satellite-derived coastline map to constrain translational position and heading. These measurements are fused within a factor graph framework to achieve globally consistent localization.

\subsubsection{LiDAR-Based Ground Plane Estimation on Water}

\paragraph{Physical Characteristics of the Water-Laser Interface}

Extracting a reference ground plane from a marine vehicle differs fundamentally from terrestrial ground extraction. On land, the ground provides a dense, continuous, and highly reflective return, representing a major portion of the LiDAR scan within close proximity of the sensor~\cite{groundgrid}. Conversely, the electromagnetic properties of water at near-infrared wavelengths lead to complex return dynamics. For most water returns, the laser beam is either absorbed or reflected away from the sensor, resulting in \say{empty} (zero-range) returns. At the crests of breaking waves or within the vehicle's wake, aerated water bubbles create diffuse scattering, producing highly localized, noisy, and spurious returns \cite{lidar_aeration}. On calm, glass-like water surfaces, the water can act as a mirror, producing specular reflections of overhead terrestrial structures rather than the water surface itself. 
While operating in coastal waters, the densest returns typically originate from nearby landmasses and vegetation. However, these structures occupy only a small fraction of the overall field of view because of the vehicle's operating distance from shore.

\paragraph{Range Image Processing Pipeline}

\begin{figure}[t]
    \centering
    \includegraphics[width=\columnwidth]{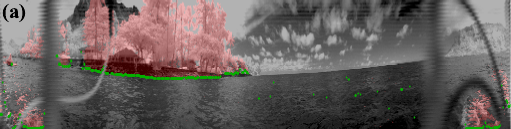}
    \vspace{0.5mm}
    \includegraphics[width=\columnwidth]{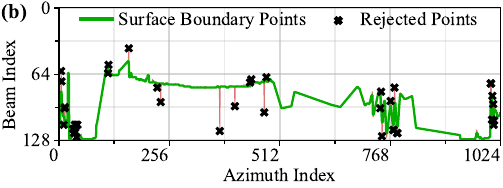}
    \vspace{0.5mm}
    \includegraphics[width=\columnwidth]{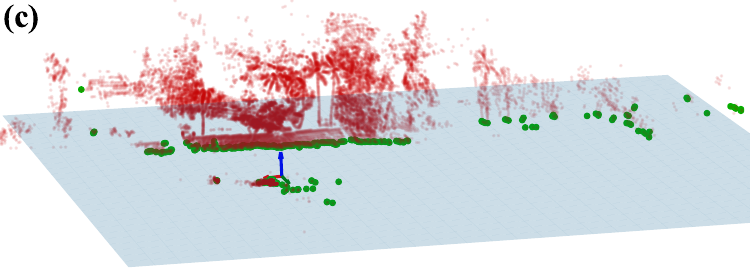}
    \caption{
    LiDAR-based water-plane extraction pipeline. (a) Morphologically filtered range image projected onto the corresponding near-infrared image, with the extracted first-return boundary highlighted in green. (b) Surface-boundary extraction in range-image space, showing rejected detections (black) and the filtered boundary signal (green). (c) RANSAC plane fit to the extracted boundary points (green) within the full LiDAR scan (red). The estimated plane is used to derive absolute roll, pitch, and heave measurements relative to the local water surface.
    }
    \label{fig:water_plane_estimation}
\end{figure}

To robustly isolate the water surface plane from these noisy return dynamics, we leverage the structured projection of the LiDAR range image. Each row corresponds to a LiDAR beam index, the horizontal axis corresponds to the azimuth sampling index, and the value at each pixel denotes the measured range.

The complete extraction pipeline is illustrated in Fig.~\ref{fig:water_plane_estimation} and proceeds as follows. First, a morphological closing operation (erosion followed by dilation) is applied to the range image to suppress small isolated returns. 
Next, for each horizontal column, a bottom-up search begins at the bottom row (the lowest depression angle) and proceeds upward to find the first valid non-zero range return. This boundary corresponds to the physical transition where the LiDAR first encounters a reflecting surface, typically the shoreline, the vehicle's wake, or larger wave crest returns. The extracted boundary is smoothed using a one-dimensional median filter applied along the horizontal image axis. 
The smoothed boundary is then back-projected into 3D space through the corresponding LiDAR measurements, yielding a set of shoreline and wake points from which a best-fit plane is estimated using RANSAC.

\paragraph{Extrapolating Attitude and Heave}

Because we define the local water plane as the global ground reference, the extracted plane parameters map the world-frame vertical axis into the LiDAR sensor frame. The relationship between this world-frame vertical and the sensor-frame normal is governed by the vehicle's orientation. This allows us to extract the absolute roll ($\phi$), pitch ($\theta$), and heave ($z$) using the following equations:
$$
\begin{aligned}
\phi &= \arctan2(n_y, n_z) 
\\ \theta &= \arcsin(-n_x) 
\\z &= d \cos\phi \cos\theta
\end{aligned}
$$
where $n_x, n_y, n_z$ are the estimated plane normal values and $d$ is the orthogonal distance.

\subsubsection{LiDAR Cross-View Satellite Image Alignment}

While the extracted shoreline points could theoretically be matched directly to a shoreline map derived from overhead satellite imagery, overhanging vegetation such as tree canopies frequently obscure the true waterline from an overhead perspective. As a result, direct matching between LiDAR shoreline observations and satellite-derived coastlines can be highly inaccurate. To overcome this limitation, we generate an orthographic projection of the visible shoreline that approximates the satellite viewpoint.

\paragraph{Simulated Satellite View Generation}

To create a simulated satellite view, we first correct the raw 3D LiDAR point cloud using the roll and pitch calculated from our plane estimation pipeline, projecting the points into a gravity-aligned local frame. We then collapse the vertical coordinate of the rectified points, projecting them orthographically onto a 2D plane to construct a binary occupancy grid with a cell resolution of 1~m. To isolate the true structural boundary, we perform morphological and spatial filtering on the grid. 
First, we remove the wake using a specified radial gate threshold within the vehicle's frame, as it is not a persistent landmark and cannot be aligned with a satellite image.
Next, morphological filtering is applied to erode noisy returns and dilate the remaining points to fill in gaps in the sparse canopy. We extract the exterior contour of the filtered occupancy grid and cast radial rays outward from the sensor frame. Along each ray, we select only the first intersecting contour point. This operation filters out deep inland terrain returns and isolates the immediate visible shoreline, directly matching the perspective of an overhead satellite image.

\paragraph{Global Registration via ICP Optimization}

An orthographic prior map of the coastline is generated through segmentation of georeferenced satellite imagery. As this map can be prepared \textit{a priori}, we create it via manual labeling. While this process can be automated using existing geo-spatial water segmentation techniques \cite{8286914, WIELAND2023113452, VOS2019104528}, automated segmentation remains beyond the scope of this work.

To register our simulated satellite view points to this prior map, we construct a localized ICP optimization. Registration is initialized from the current pose estimate to restrict the search to a local spatial window around the vehicle and to prevent convergence to local minima in repetitive environments. The optimization is constrained to the three planar degrees of freedom: position $(x, y)$ and heading $(\psi)$.

\subsubsection{Factor Graph SLAM with LiDAR Odometry}

To estimate the trajectory of the surface vessel, the proposed LiDAR framework fuses relative motion estimates with absolute observations derived from the coastal environment within a factor graph formulation. The resulting fixed-lag smoother, implemented in GTSAM~\cite{gtsam} and optimized incrementally using iSAM2~\cite{isam2}, combines LiDAR odometry with water-surface and shoreline-based measurements to produce globally referenced pose estimates.

\begin{figure}[b]
    \centering
    \includegraphics[width=\columnwidth]{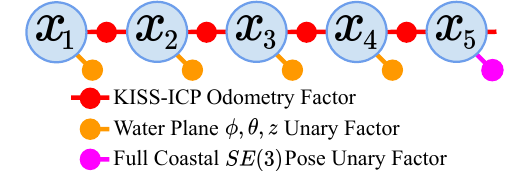}
    \caption{LiDAR fixed-lag factor graph architecture. Consecutive vehicle poses are connected by KISS-ICP odometry factors. Water-Surface factors provide roll, pitch, and height observations at each scan. Every tenth scan, these observations are fused with cross-view shoreline registration measurements to form a globally referenced Coastal Pose factor that constrains the full $SE(3)$ state.}
    \label{fig:lidar_fg}
\end{figure}

The local odometry backbone uses KISS-ICP~\cite{vizzo2023kiss}, which estimates relative rigid-body transformations between consecutive LiDAR scans. These transformations are represented as binary odometry factors within the graph. Our choice of KISS-ICP is motivated by two primary considerations. First, KISS-ICP consistently outperforms alternative direct and feature-based LiDAR odometry pipelines in sparse, feature-light coastal scenes, as demonstrated by the empirical evaluation in Table~\ref{tab:lidar_trajectory_comparison}. Second, its clean, modular architecture allows straightforward extraction and integration of relative pose measurements into a factor graph. 

\begin{figure*}[h]
    \centering
    \includegraphics[width=\textwidth]{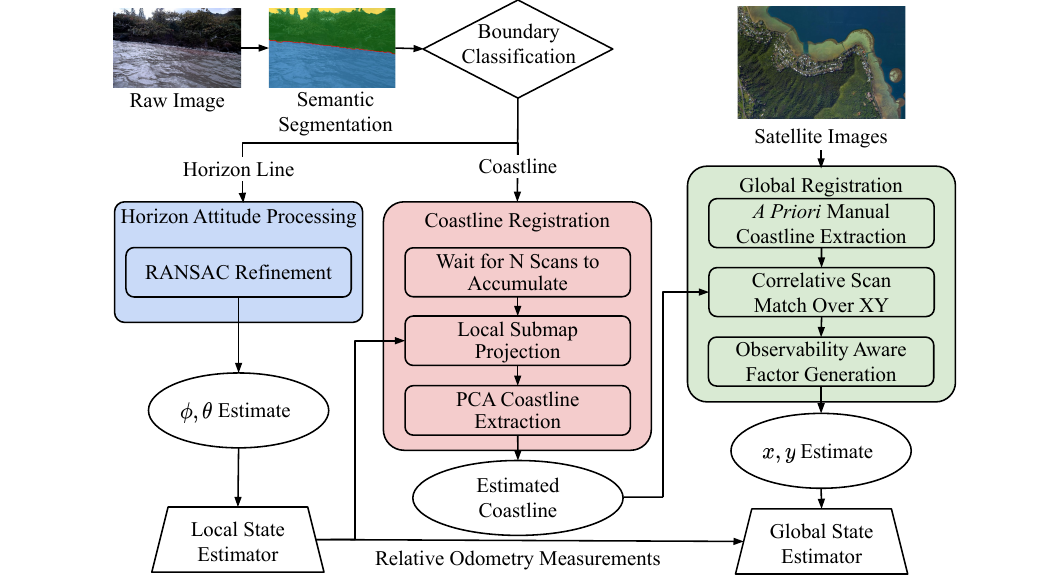}
    \caption{Architecture of the visual coastline localization pipeline. The system branches into two processes based on the semantic segmentation of the input image. If the upper water boundary is classified as sky (horizon line), the vehicle's roll~($\phi$) and pitch~($\theta$) are estimated and a partial attitude measurement is integrated into the local state estimator. Conversely, if the boundary is classified as a coastline, sequential observations are projected and coalesced into a local submap based on the current odometry estimate. This local submap is then aligned against an \textit{a priori} satellite map via correlative scan matching. Because visual coastline projections lack the fidelity of LiDAR, heading is assumed observable via a magnetometer, restricting the correlative scan match to an exhaustive search over a discretized $XY$ translation window around the prior state estimate. This exhaustive search generates an observability-aware factor that provides a strong geometric constraint on cross-shore distance while remaining loosely constrained along the shore. 
    Finally, the resulting coastline registration factor and the relative odometry estimates from the local state estimator are incorporated into the global state estimator.}
    \label{fig:visual_pipeline}
\end{figure*}

Although KISS-ICP provides reliable short-term motion estimates, drift inevitably accumulates over extended distances. To bound this drift, we augment KISS-ICP with observations from the water-surface ground plane and LiDAR cross-view satellite alignment, forming the proposed \say{Coastal-KISS} framework. The water-plane estimate constrains roll $(\phi)$, pitch $(\theta)$, and height $(z)$, and shoreline registration constrains planar position $(x,y)$ and yaw $(\psi)$, as illustrated in Fig.~\ref{fig:lidar_fg}.

The estimated water plane is incorporated as a 10~Hz unary factor. Every tenth scan, the water-plane and shoreline registration measurements are jointly incorporated as a full $SE(3)$ unary pose factor. Together, these absolute observations bound drift accumulated by the relative odometry factors while maintaining real-time state estimation.

\subsection{Vision-Based Localization}
\label{subsec:vision_localization}
The proposed visual localization pipeline extracts both horizon and shoreline observations from semantically labeled monocular imagery. Horizon observations provide roll and pitch constraints, while reconstructed shoreline observations are aligned against a satellite-derived coastline map to recover globally referenced position measurements. As illustrated in Fig.~\ref{fig:visual_pipeline}, the framework consists of five primary stages: semantic segmentation, horizon-based attitude estimation, shoreline reconstruction, satellite-based registration, and incorporation of the resulting observations within a hierarchical global estimator.

\subsubsection{Semantic Segmentation and Boundary Detection}

Prior to estimating the physical location of the coastline, the coastline and horizon must first be detected in image space. This is achieved through semantic segmentation, where each pixel is assigned one of four classes: water, sky, land, or background. The coastline is defined as the water-land boundary, while the horizon is defined as the water-sky boundary.

To ensure a generalizable solution that avoids hand-labeled training data, we employ a zero-shot foundation-model pipeline to partition each pixel into these semantic classes. Our perception pipeline combines Grounding DINO~\cite{liu2024grounding} with the Segment Anything Model 2 (SAM~2)~\cite{ravi2025sam}. Because Grounding DINO is a large transformer-based model, executing it on every frame would be prohibitively expensive for real-time operation. Instead, Grounding DINO is used only to initialize semantic prompts, after which SAM~2 segments and tracks the corresponding regions across sequential frames using its predictive memory model.

Although this eliminates the need to repeatedly execute Grounding DINO, SAM~2 remains computationally demanding, particularly in a multi-camera configuration. To maintain real-time operation, the 15~Hz camera streams are downsampled by processing only one out of every four frames, resulting in an effective update rate of 3.75~Hz. Even with the SAM~2 predictive memory model, segmentation masks are still prone to drift, and masks may be lost, if for example close proximity to shore obscures the sky. To account for this as well as to mitigate VRAM accumulation when tracking long sequences of images, Grounding DINO re-initializes the SAM~2 masks every 100 frames.

To locate the coastline or horizon line in a semantic image, a boundary extraction method is performed via a column-wise bottom-up search to locate the boundary transitions. For each column, the algorithm extracts a single coastline or horizon point in pixel coordinates where the boundary is the median pixel location between the top of the water mask and the bottom of either the land or sky mask. Semantic boundaries, while accurate on an image level scale, can still be noisy at the pixel level scale and may not provide a crisp boundary edge. Because the horizon should represent a straight line, the undistorted, extracted horizon boundary points are filtered using RANSAC before being passed to an attitude observer to estimate the roll and pitch of the vehicle using the methods presented in~\cite{marine_vessel_attitude}. Coastline boundary points are saved in memory to be used later in constructing submaps.

\begin{figure}
    \centering
    \includegraphics[width=\columnwidth]{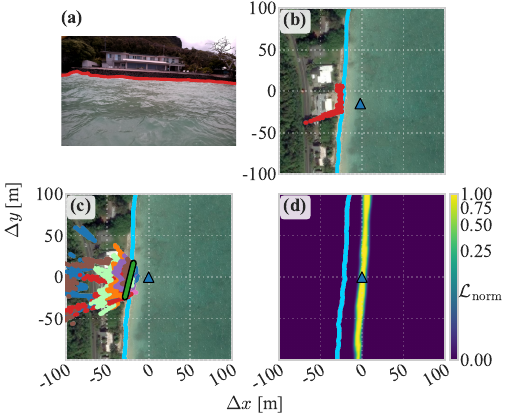}
    \caption{Visual coastline localization pipeline. (a) Extracted coastline observation (red) overlaid on the input image. (b) Single-image shoreline projection from image space into geospatial coordinates, overlaid on the satellite coastline map (light blue); the vehicle position is shown as a blue triangle. (c) Accumulated shoreline observations (multiple colors) forming a 10-second shoreline submap, with the dominant shoreline trend estimated using PCA (green line). The persistent breaking waves visible along the left side of the image introduce a local distortion in the projected shoreline that remain evident in the extracted PCA line. (d) Correlative scan-matching likelihood generated by registering the PCA shoreline estimate against the satellite coastline map. The resulting likelihood distribution strongly constrains cross-shore position while remaining weakly constrained along the shoreline.
    }
    \label{fig:visual_segment}
\end{figure}

\subsubsection{Local Submap Construction}

Once the coastline has been identified in image space, the next step is to recover its physical location relative to the vehicle, as illustrated in Fig.~\ref{fig:visual_segment}. Unlike LiDAR, which directly measures range, monocular cameras must infer shoreline distance through geometric reconstruction. 
Under the assumptions of the proposed coastal scene model, coastline observations lie on the local water surface, and the camera height above this surface, $h_{\text{camera}}$, is known. 
Each shoreline pixel defines a viewing ray originating from the camera center. Using the estimated vehicle roll and pitch, this ray is transformed into the world frame and intersected with the mean water surface, yielding the corresponding shoreline location.

For the principal viewing plane illustrated in Fig.~\ref{fig:coastline_assumptions}, this reconstruction simplifies to 
$$
d_{\text{shore}}=h_{\text{camera}}\tan(\alpha),
$$
where $\alpha$ is the declination angle of the viewing ray expressed in the world frame. 
The corresponding reconstruction for arbitrary image pixels is described by Kiefer et al.~\cite{10342453}.

While shoreline reconstruction is mathematically straightforward under the proposed coastal scene assumptions, individual shoreline projections provide only noisy instantaneous measurements. 
Small attitude errors and uncertainty in the shoreline boundary produce large range errors as distance to shore increases.
To suppress these errors, sequential shoreline observations are accumulated into a local submap and filtered prior to satellite registration.

Constructing a stable shoreline submap requires expressing coastline observations collected over several seconds into a common reference frame. Unlike LiDAR submaps, which can often be formed from dense scans over fractions of a second, visual shoreline submaps require accurate ego-motion estimates over a 10-second accumulation window, exceeding what an IMU can reliably track on its own. To obtain this estimate, we employ a high-frequency fixed-lag factor graph operating in a local odometry frame. The estimator fuses pre-integrated IMU factors~\cite{imu_preintegration}, with magnetometer heading, horizon-derived roll and pitch observations, a weak forward-velocity prior based on the commanded vehicle speed, and a weak altitude prior centered on the camera height above the water surface. 
The resulting trajectory is used to project all coastline observations into a common coordinate system before 
satellite registration.

To simplify the submap representation prior to alignment, all coastline observations are projected into a 2D histogram grid with 1~m resolution. As an initial filtering step, cells falling below the 90th percentile of point density are discarded. 
While the remaining observations are still noisy, the visible coastline is generally well approximated by a locally linear segment over a 10-second observation window. Principal Component Analysis (PCA) is applied to estimate the dominant shoreline orientation, producing a representative line for subsequent satellite scan matching.

\subsubsection{Satellite Registration via Correlative Matching}

\begin{figure*}[h]
    \centering
    \includegraphics[width=\textwidth]{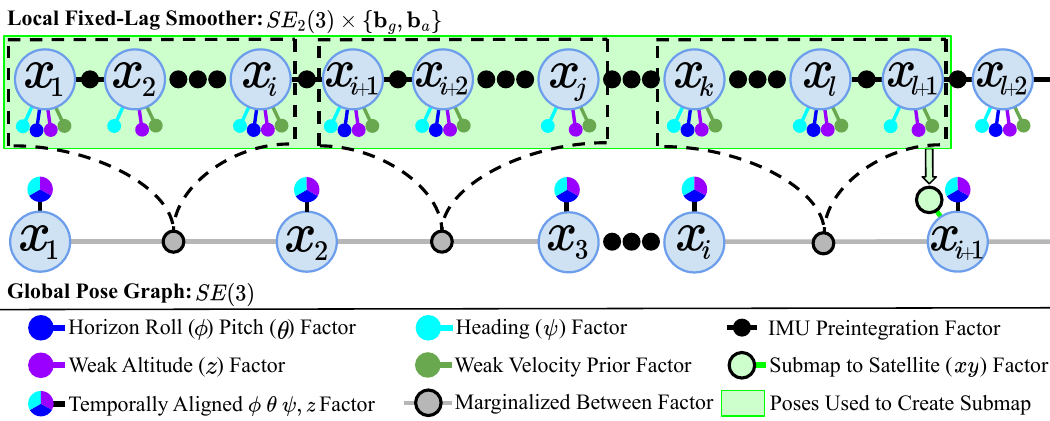}
    \caption{Hierarchical factor graph architecture for the visual localization pipeline. The local fixed-lag smoother estimates the high-rate $SE_2(3)\times \text{biases}$ state within a local odometry frame using IMU pre-integration together with horizon, magnetometer, and weak altitude constraints. Marginalized relative pose factors connect the local estimator to a sparse global $SE(3)$ pose graph, where temporally aligned orientation observations and a weak altitude prior form unified unary pose constraints. Local trajectory estimates are accumulated into shoreline submaps, which are registered against satellite imagery to provide global $x,y$ position constraints.}
    \label{fig:visual_fg}
\end{figure*}

While the local submap provides a consistent representation of the visible coastline, it remains expressed in the local odometry frame. To recover the vessel's global position, this representation must be registered against a globally referenced coastline map. Using the same satellite-derived coastline map as the LiDAR pipeline, a $200\text{ m} \times 200\text{ m}$ sub-grid centered on the vehicle's current position estimate is extracted from the global coastline map. The PCA shoreline representation is then registered against this local satellite map region using a 2D correlative scan matching approach~\cite{5152375}. Unlike the LiDAR submaps, which contain rich two-dimensional structure, the vision-based shoreline representation is typically dominated by a single linear feature. Estimating yaw from such observations introduces severe rotational ambiguity and unreliable registrations. We therefore assume yaw is independently observed by the magnetometer and restrict the correlative search to planar $(x,y)$ translations.

Even when the search is restricted to translation, registering a linear coastline feature against a locally linear satellite map boundary remains geometrically degenerate along the shoreline's longitudinal axis, while remaining tightly constrained in the perpendicular direction. To quantify this directional observability, the correlative scan matcher produces a registration heatmap representing the alignment score over the search space. PCA is applied to the high-scoring region of this heatmap to estimate its principal directions and corresponding eigenvalues. These quantities characterize the uncertainty of the registration and are used to construct a unary factor with directionally varying uncertainty. As a result, the factor provides a stronger constraint perpendicular to the coastline than along it.

\subsubsection{Hierarchical Factor Graph Optimization}
\label{subsubsec:vision_hierarchy}
As the proposed registration primarily constrains motion perpendicular to the shoreline, meaningful localization updates occur only when the coastline exhibits sufficient geometric variation. Consequently, long, straight sections of coastline present a classic SLAM dilemma commonly referred to as the \say{hallway problem}. The distance to the shoreline remains tightly constrained, while the position along the shoreline drifts owing to the limited observability of the registration. 
As a result, substantial longitudinal drift may accumulate before the coastline eventually changes direction, at which point the estimator must retroactively distribute the accumulated corrections along the preceding trajectory.

However, relying on a standard single-estimator framework to handle this backtracking introduces a critical trade-off: a standard fixed-lag smoother marginalizes older states to maintain real-time performance, preventing retroactive updates, while a monolithic full-history factor graph quickly becomes computationally intractable over long-duration runs.

To overcome this trade-off, we decouple local trajectory estimation from global optimization through the hierarchical factor graph architecture shown in Fig.~\ref{fig:visual_fg}. 
The local odometry layer employs the same fixed-lag smoother used to project coastal observations into a common local frame for submap construction. The global optimization layer maintains a sparse full-history pose graph that estimates the complete vehicle trajectory while incorporating intermittent submap alignment measurements. Both layers are implemented in GTSAM~\cite{gtsam} and optimized incrementally using iSAM2~\cite{isam2}, enabling online inference while preserving the ability to incorporate delayed global corrections.

Keyframe nodes are appended to the global graph at 1~Hz and connected by binary relative pose factors extracted from the marginalized state and covariance of the local smoother. This decoupled architecture enables the local fixed-lag smoother to provide high-rate motion estimates for submap construction while the global graph preserves the full trajectory history required to absorb delayed coastline constraints. 
In turn, when the vessel encounters an informative section of coastline, the global optimizer can distribute the resulting correction across the preceding trajectory while maintaining real-time operation.

Temporally aligned magnetometer, horizon attitude, and weak height observations are combined into a single unary factor at each global keyframe. Horizon observations are not always available due to dropped frames, timestamp misalignment, or an unobserved horizon. While missing horizon measurements simply omit the corresponding constraint from the local estimator, the global unary factor substitutes missing roll and pitch observations with the corresponding estimates from the local fixed-lag smoother.

\section{RESULTS AND INSIGHTS}
\label{sec:results}
This section evaluates the proposed localization frameworks across three real-world coastal datasets. We first introduce the experimental platform and datasets before evaluating the LiDAR and monocular localization pipelines independently. 
For the vision pipeline, we additionally evaluate the perception front-end and conclude with an ablation study examining why conventional LiDAR odometry methods often struggle in coastal environments.

\subsection{System and Dataset Overview}
Experimental data was collected using a retrofitted WAM-V 8 equipped with three AR0234 2.3~Mpix global-shutter color cameras and an Ouster OS1-128 LiDAR. Inertial measurements were provided by an SBG Ellipse-D IMU, while dual-antenna RTK-GPS provided high accuracy position and heading measurements as a quantitative evaluation. The IMU, LiDAR, and onboard computers were synchronized using Precision Time Protocol (PTP, IEEE 1588), while camera timestamps were synchronized through the manufacturer's hardware timing interface. Intrinsic calibration and camera--IMU extrinsics were estimated using the Kalibr toolkit~\cite{kalibr1,kalibr2}, while the LiDAR--IMU transformation was obtained from the sensor platform CAD model.

Establishing centimeter-level ground truth over kilometer-scale marine trajectories is challenging, as traditional survey techniques requiring continuous line-of-sight are impractical for missions of this duration. To obtain a high-fidelity reference solution, we employ a fixed-lag factor graph disciplined by dual-antenna RTK-GPS as our reference trajectory. 
A new pose node is added for each RTK-GPS update (5~Hz), while pre-integrated IMU measurements form binary factors connecting consecutive pose nodes.
The RTK-GPS reports a positional standard deviation of 1.4~cm, while dual-antenna heading is specified with a standard deviation of 0.34$^\circ$. When available, horizon observations constrain roll and pitch with an estimated standard deviation of approximately 0.2$^\circ$~\cite{benham2026invariantextendedkalmanfilter}.
Since the positional errors evaluated in this work are on the order of meters, the resulting reference trajectory provides sufficient accuracy for quantitative comparison.

\begin{figure}
    \centering
    \includegraphics[width=\columnwidth]{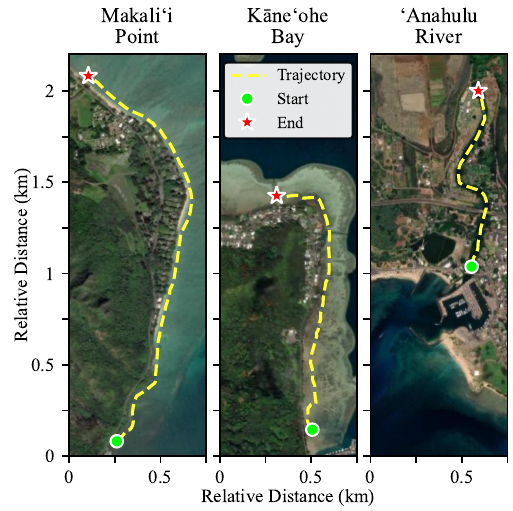}
    \caption{Overhead view of the three dataset trajectories collected in O‘ahu, Hawaii. Imagery and trajectories are rotated vertically to create a concise visual display with a consistent scale. Axes represent relative local distance in kilometers, with start (green circle) and end (red star) points marked for each trajectory (yellow).}
    \label{fig:dataset}
\end{figure}

\begin{table}[htbp]
\caption{Dataset characteristics and environmental scale}
\label{tab:dataset_overview}
\centering
\setlength{\tabcolsep}{3.5pt} 
\footnotesize %
\begin{tabular}{lccccc}
\toprule
Dataset Name & 
\begin{tabular}[c]{@{}c@{}}Length\\ (km)\end{tabular} & 
\begin{tabular}[c]{@{}c@{}}Time\\ (min)\end{tabular} & 
\begin{tabular}[c]{@{}c@{}}Velocity\\ (m/s)\end{tabular} & 
\begin{tabular}[c]{@{}c@{}}Mean Dist.\\ Shore (m)\end{tabular} & 
\begin{tabular}[c]{@{}c@{}}Max Dist.\\ Shore (m)\end{tabular} \\
\midrule
Makali\textquoteleft{}i Point  & 2.3 & 18 & 2.2 & 25.0 & 48 \\
K\={a}ne\textquoteleft{}ohe Bay & 1.5 & 15 & 1.8 & 41.0 & 67 \\
\textquoteleft{}Anahulu River  & 1.1 & 16 & 1.2 &  7.5 & 16 \\ 
\end{tabular}
\end{table}

The vehicle was deployed around the north shore of O\textquoteleft{}ahu, Hawai\textquoteleft{}i, in August 2025, where three coastal datasets were collected. Together, these datasets span diverse shoreline structure, vegetation densities, stand-off distances, and sea states, ranging from calm inland waters to exposed coastal conditions. Throughout each mission, the vehicle maintained a safe stand-off distance while operating approximately parallel to the shoreline. Representative trajectories are shown in Fig.~\ref{fig:dataset}, while quantitative dataset characteristics are summarized in Table~\ref{tab:dataset_overview}. Together, these datasets provide a representative benchmark for evaluating both localization pipelines under diverse coastal operating conditions.

\subsection{LiDAR Results}
We begin by evaluating the performance of modern LiDAR odometry pipelines in coastal environments to establish the challenges posed by maritime operation. We then demonstrate how exploiting coastal scene geometry enables accurate estimation of roll, pitch, and heave from the water surface, before showing how these observations, together with shoreline-based satellite alignment, improve long-term localization accuracy.

\subsubsection{LiDAR Pipeline Evaluation on Coastal Data}
\begin{table}[b]
\centering
\footnotesize %
\setlength{\tabcolsep}{3.0pt} %
\caption{LiDAR odometry comparison. Metrics are presented as translation [m] / rotation [deg] (ATE RMSE). Bold denotes best results, and `--' indicates failure to produce a solution.}
\label{tab:lidar_trajectory_comparison}
\begin{tabular}{l cc cc cc cc cc}
\toprule
\multirow{2}{*}{Dataset} & \multicolumn{2}{c}{KISS-ICP} & \multicolumn{2}{c}{FORM} & \multicolumn{2}{c}{LOAM} & \multicolumn{2}{c}{LIO-SAM} & \multicolumn{2}{c}{GenZ-ICP} \\
\cmidrule(lr){2-3} \cmidrule(lr){4-5} \cmidrule(lr){6-7} \cmidrule(lr){8-9} \cmidrule(lr){10-11}
 & $T$ & $R$ & $T$ & $R$ & $T$ & $R$ & $T$ & $R$ & $T$ & $R$ \\
\midrule
Makali\textquoteleft{}i Point & \textbf{42.0} & \textbf{2.6} & 80 & 11 & 123 & 15 & 844 & 56 & -- & -- \\
K\={a}ne\textquoteleft{}ohe Bay   & \textbf{22.4} & \textbf{3.0} & 58 & 14 & 1153 & 120 & 148 & 17 & 61 & 41 \\
\textquoteleft{}Anahulu River & 5.1           & \textbf{0.8}          & \textbf{4.1} & \textbf{0.8} & 20  & 4.2  & 6.1   & 1.4 & -- & -- \\
\bottomrule
\end{tabular}
\end{table}
Recent LiDAR odometry research has increasingly focused on developing generalized pipelines capable of operating across a wide range of environments. However, most benchmarking datasets are collected in terrestrial settings with dense returns and a consistent ground plane. Coastal environments present substantially different sensing conditions, motivating an evaluation of how well these general-purpose terrestrial methods transfer to maritime operation.

We evaluated both seminal and recent LiDAR odometry pipelines using the Evalio benchmarking framework~\cite{evalio}. The methods include KISS-ICP~\cite{vizzo2023kiss}, FORM~\cite{potokar2025formfixedlagodometryreparative}, LOAM~\cite{loam}, LIO-SAM~\cite{liosam}, and GenZ-ICP~\cite{lee2024genzicp}. To ensure a fair comparison, the primary tunable parameters for each pipeline were optimized through parameter sweeps within Evalio, with each method evaluated using its best-performing configuration.

As shown in Table~\ref{tab:lidar_trajectory_comparison}, most evaluated LiDAR odometry pipelines struggled in the open water coastal datasets, with several failing to produce a valid solution altogether. Although the precise failure mechanisms differ between methods, we attribute the overall degradation to the unique characteristics of coastal environments, including the absence of a stable ground plane, complex water-surface returns, aggressive wave-induced motion, and the limited availability of dense, persistent LiDAR returns. 
Interestingly, the water-return ablation study presented later suggests that water returns themselves are not the dominant failure mode, indicating that the scarcity of stable LiDAR observations may play a more significant role. Among the evaluated methods, KISS-ICP consistently achieved the best performance in the open-water datasets, while FORM performed best in the sheltered, feature-rich river environment.

\subsubsection{Ground Plane Estimation Accuracy}
Since the proposed localization framework relies on the water surface as a geometric reference, we first evaluate the accuracy of the LiDAR front-end responsible for estimating the water plane. 
Roll, pitch, and heave are recovered independently from each LiDAR scan, and Table~\ref{tab:coastline_observer_errors} reports the mean and standard deviation of the resulting errors across all scans in each dataset.

On the two open-water datasets, the proposed method estimates roll and pitch with mean errors below 0.2$^\circ$ and standard deviations below 1$^\circ$. Likewise, the recovered heave remains accurate to within 4~cm on average, with less than 1~cm standard deviation.

Performance degrades slightly in the sheltered river environment due to an unexpected failure mode of the front-end water-plane extraction. 
Unlike in the open-water datasets, the calm water surface frequently produces specular reflections that cause above-water objects to be mirrored beneath the water plane.
These reflections violate the assumed water-plane geometry and distort the resulting plane estimate. Because the mirrored returns are geometrically consistent with valid shoreline observations within an individual LiDAR scan, the proposed range-image filtering pipeline has insufficient information to distinguish them reliably. 
Incorporating a coarse water-plane estimate from the previous state could help identify and reject these reflections before plane fitting. However, when the water plane is estimated from a single LiDAR scan without prior state information, this ambiguity remains.

\begin{table}[t]
\caption{LiDAR ground plane estimation accuracy across environments}
\label{tab:coastline_observer_errors}
\centering
\setlength{\tabcolsep}{3.5pt} 
\footnotesize %
\begin{tabular}{lccc}
\toprule
Mission Name & 
\begin{tabular}[c]{@{}c@{}}Roll Error\\ (deg)\end{tabular} & 
\begin{tabular}[c]{@{}c@{}}Pitch Error\\ (deg)\end{tabular} & 
\begin{tabular}[c]{@{}c@{}}Z Error\\ (m)\end{tabular} \\ 
\midrule
Makali\textquoteleft{}i Point                     & $-0.11 \pm 0.86$ & $\phantom{-}0.02 \pm 0.67$  & $-0.04 \pm 0.09$ \\
K\={a}ne\textquoteleft{}ohe Bay & $-0.02 \pm 0.56$ & $-0.18 \pm 0.28$ & $-0.01 \pm 0.06$ \\
\textquoteleft{}Anahulu River                  & $\phantom{-}0.04 \pm 1.86$  & $-0.61 \pm 1.07$ & $\phantom{-}0.41 \pm 0.39$ \\ 
\bottomrule
\end{tabular}
\end{table}

\subsubsection{Coastal-KISS Trajectory Accuracy}

\begin{figure}[h]
    \centering
    \includegraphics[width=\columnwidth]{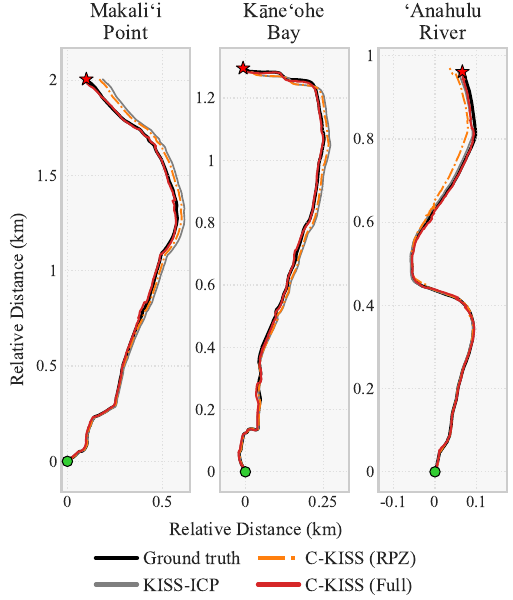}
    \caption{
    Comparison of trajectories produced by the proposed Coastal-KISS pipeline and KISS-ICP baseline across the three evaluation datasets.
    }
    \label{fig:trajectory_comp_lidar}
\end{figure}

Having validated the accuracy of the proposed water-plane observations, we now evaluate their impact on long-term trajectory estimation. 
The first incorporates water-plane observations alone, while the second additionally incorporates shoreline-to-satellite registration to create full pose observations. The resulting trajectories are shown in Fig.~\ref{fig:trajectory_comp_lidar}, with quantitative accuracy summarized in Table~\ref{tab:coastal_kiss}.

The results highlight that KISS-ICP, like many LiDAR odometry pipelines, accumulates drift in both the horizontal 
and vertical 
directions.

Using only the water-plane observations, Coastal-KISS substantially reduces attitude and vertical translation error in the two open-water datasets, confirming that the extracted water-plane measurements provide effective roll, pitch, and height constraints. As expected, improvements in horizontal position remain modest because no absolute planar position information is introduced. In the river dataset, however, the baseline KISS-ICP solution achieves lower rotation and overall position error, reflecting the degradation of the water-plane estimate under calm-water reflections.
When shoreline-to-satellite registration is incorporated, horizontal drift is significantly reduced, yielding the most accurate overall trajectories in the open-water datasets and matching the best-performing baseline in the river environment.

\begin{table}[t]
\centering
\footnotesize 
\setlength{\tabcolsep}{3.0pt} %
\caption{Trajectory comparison of the proposed Coastal-KISS framework. Metrics are reported as translation $T_{xy}$, $T_z$ [m] and rotation $R$ [deg] (ATE RMSE).}
\label{tab:coastal_kiss}
\begin{tabular}{l ccc ccc ccc}
\toprule
\multirow{2}{*}{Dataset} & \multicolumn{3}{p{1.9cm}}{\centering Coastal-KISS \\ w/ Satellite} & \multicolumn{3}{p{1.9cm}}{\centering Coastal-KISS \\ (Water Plane)} & \multicolumn{3}{p{1.9cm}}{\centering KISS-ICP \\ \vspace{0pt}} \\
\cmidrule(lr){2-4} \cmidrule(lr){5-7} \cmidrule(lr){8-10}
 & $T_{xy}$ & $T_z$ & $R$ & $T_{xy}$ & $T_z$ & $R$ & $T_{xy}$ & $T_z$ & $R$ \\
\midrule
Makali\textquoteleft{}i Point & \textbf{7.2} & 0.14 & 2.4 & 24.8 & \textbf{0.13} & \textbf{2.1} & 40.3 & 12.2 & 2.6 \\
K\={a}ne\textquoteleft{}ohe Bay   & \textbf{8.1} & 0.07 & 3.2 & 20.4 & \textbf{0.06} & \textbf{2.9} & 22.4 & 0.8  & 3.0 \\
\textquoteleft{}Anahulu River & \textbf{4.5} & \textbf{0.6}  & 3.3 & 16.5 & \textbf{0.6}  & 5.1 & 4.8  & 1.8  & \textbf{0.8} \\
\bottomrule
\end{tabular}
\end{table}

\begin{table*}
\centering
\caption{Comparison of Semantic Segmentation and Shoreline Extraction Performance Across Vision Architectures. Bold and underlined values indicate the best and second-best results, respectively.}
\label{tab:segmentation_comparison}
\begin{tabular}{l ccccccc}
\toprule
Model & mIoU & Water IoU & Sky IoU & Land IoU & Shoreline 95\% Error (px) & Shoreline Failure & FPS \\
\midrule
Grounded-SAM 2        & \textbf{0.95} & \textbf{0.99} & \textbf{0.90} & \textbf{0.96} & 7.0  & \textbf{0\%}  & 17.5 \\
Grounded-MobileSAM   & \textbf{0.95} & \textbf{0.99} & \textbf{0.90} & \textbf{0.96} & \underline{5.7}  & \textbf{0\%}  & 5.2  \\
YOLOE                & 0.88 & 0.98 & 0.72 & 0.92 & 9.0  & \textbf{0\%}  & \underline{117} \\
YoloWorld-SAM 2      & 0.64 & 0.85 & 0.35 & 0.73 & 6.4  & 13\% & 6.8  \\
YoloWorld-MobileSAM  & 0.66 & 0.86 & 0.36 & 0.76 & \textbf{5.2}  & 13\% & 17.5 \\
SAM 3                & 0.89 & 0.97 & \textbf{0.90} & 0.81 & 9.3  & \textbf{0\%}  & 6.1  \\
YOLO-seg (supervised)& 0.75 & 0.83 & 0.64 & 0.78 & 34.8 & 7\%  & \textbf{229} \\
\bottomrule
\end{tabular}
\end{table*}

\subsection{Semantic Segmentation Evaluation}

\begin{figure*}
    \centering
    \includegraphics[width=\textwidth]{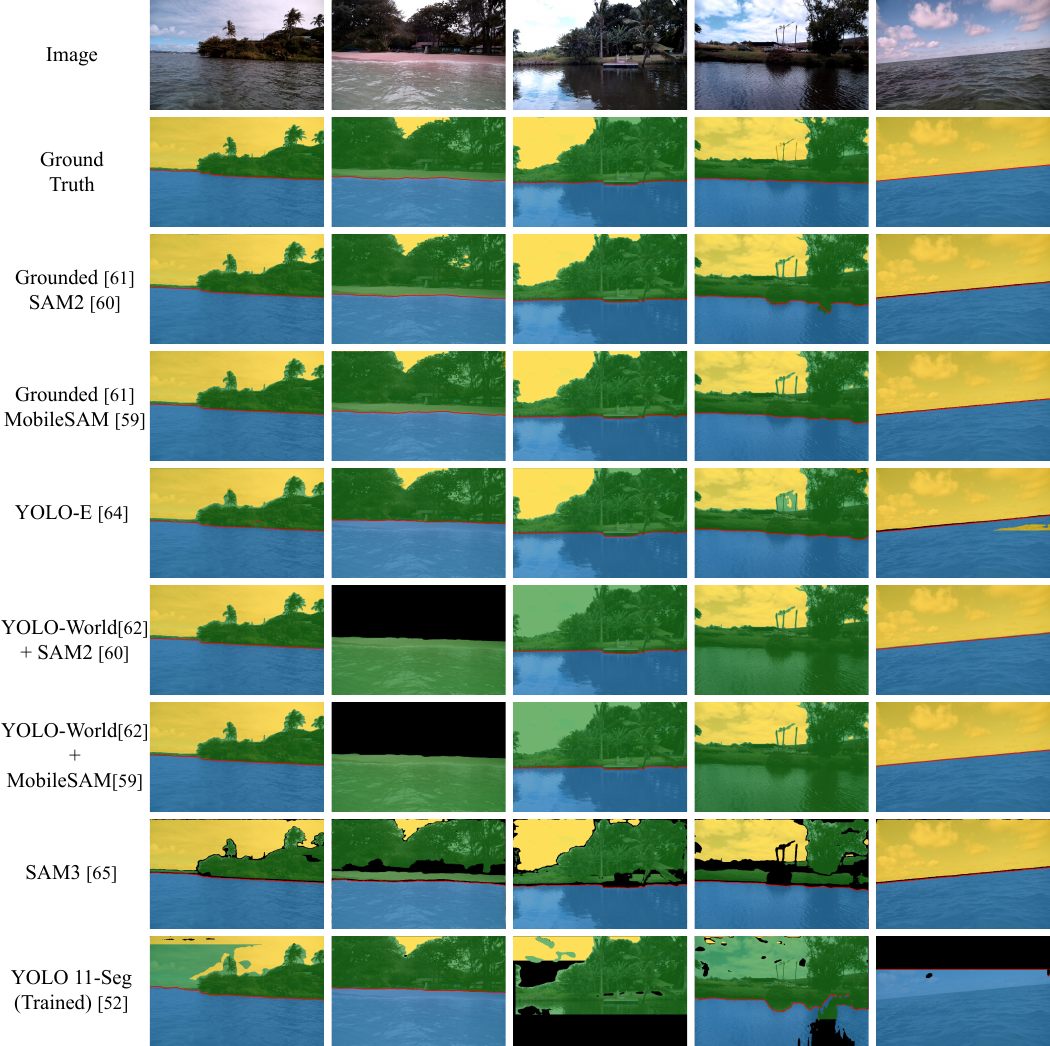}
    \caption{
    Comparison of semantic segmentation models for classifying water, sky, and land. The red curve denotes the extracted upper water boundary used for shoreline localization. Unless otherwise noted, all models are evaluated in a zero-shot setting without task-specific training. Grounded-SAM~2 and Grounded-MobileSAM produce the most consistent semantic segmentations and mask boundaries, although Grounded-SAM~2 struggles in the fourth image where calm-water reflections introduce ambiguity. YOLOE occasionally hallucinates sky within the water or misclassifies portions of the shoreline, most notably labeling the beach as water in the second image. The YOLO-World variants frequently under-segment the sky or misclassify water as land. SAM 3 produces robust segmentations but yields less precise shoreline boundaries and may omit portions of the land. 
    YOLO-seg, trained on 350 labeled images from a single location on the island, fails to generalize to new coastal environments. 
    These results suggest that transformer-based open-vocabulary detectors, such as Grounding DINO, paired with foundation segmentation models such as MobileSAM and SAM~2, provided the most reliable classification across diverse coastal scenes and were therefore selected for use in the proposed localization pipeline. 
    }
    \label{fig:model_comparison}
\end{figure*}

Fundamental to the proposed monocular localization pipeline is a reliable shoreline extractor, as all subsequent geometric reasoning depends on accurate segmentation of the water, sky, and land regions. We therefore begin by evaluating the performance of several modern segmentation approaches on our coastal dataset.

The evaluated approaches include two prompting-based foundation-model pipelines, Grounding DINO~\cite{liu2024grounding} paired with either SAM~2~\cite{ravi2025sam} (Grounded-SAM~2) or MobileSAM~\cite{zhang2023faster} (Grounded-MobileSAM), and YOLO-World~\cite{cheng2024yolo} paired with either SAM~2 or MobileSAM. We additionally evaluate SAM~3~\cite{carion2025sam}, YOLO-E~\cite{yolo_e}, and a fine-tuned YOLO11-Seg model~\cite{yolov8_ultralytics}. Together, these approaches span both foundation-model and task-specific segmentation paradigms.
Quantitative results on 30 manually labeled test images are reported in Table~\ref{tab:segmentation_comparison}, while representative qualitative examples are shown in Fig.~\ref{fig:model_comparison}.

Segmentation quality was evaluated using both region-based and shoreline-specific metrics. We report mean Intersection over Union (IoU) together with per-class IoU for water, sky, and land. However, mean IoU alone does not fully capture the segmentation accuracy required by the proposed framework, as shoreline registration depends primarily on the alignment of the water-land boundary. To directly evaluate this quantity, the upper boundary of the water mask is extracted and compared against the manually labeled shoreline. We report the 95th-percentile shoreline error in pixels, which captures the reliability of the extracted shoreline boundary under challenging conditions. Additionally, we measure failure rate, defined as images in which no valid water segmentation is produced, as well as processing speed evaluated on 100 sequential images using an NVIDIA RTX 4070.

As an initial baseline, we fine-tuned a pre-trained YOLO11-Seg model to directly classify and segment water, sky, and land regions.
The fine-tuned YOLO11-Seg model, trained on 350 labeled images from a small subsection of the Makali\textquoteleft{}i Point dataset, achieved high performance within the training environment but failed to generalize to unseen shorelines and environmental conditions. For example, later in the Makali\textquoteleft{}i Point dataset the model encounters a sandy beach that was absent from the training set and consequently misclassifies portions of the beach as water. 
This lack of generalization extends to the remaining datasets, where differences in lighting, vegetation, and shoreline appearance frequently lead to incorrect water classifications. 
To avoid developing a model that overfits to a specific coastal environment or requires extensive labeled training data, we instead sought a generalizable zero-shot foundation-model approach for shoreline segmentation.

Among the YOLO-based approaches, many variants struggled to classify land and frequently failed to classify or segment the sky, occasionally missing water entirely. This behavior is consistent with the design philosophy of YOLO-style detectors, which are primarily optimized for identifying foreground objects rather than large-scale scene regions. In many instances, the extracted water-land boundary was often incomplete, producing unreliable shoreline observations.
YOLOE generally performed better than the YOLO-World variants but still exhibited temporal consistency issues. Segmentation masks frequently varied between adjacent frames, resulting in isolated sky regions appearing within the water mask or water regions leaking into the land mask. This behavior is likely related to YOLOE treating each frame independently, as it lacks the temporal memory mechanisms available to SAM~2 and SAM~3.

Grounding DINO consistently identified the desired semantic classes across diverse environments, reflecting its strong open-vocabulary grounding capabilities. However, nearly all evaluated zero-shot segmentation pipelines except SAM~3 exhibited the same failure mode. The resulting land segmentation masks frequently corresponded only to individual vegetation clusters, rocks, or shoreline features rather than the complete land region. As a result, the extracted land masks were often fragmented and unsuitable for direct shoreline reconstruction. To address this issue, all zero-shot approaches except SAM~3 were configured to segment only the water and sky classes, with all remaining pixels assigned to land.

When observing pure water-sky horizons, several models left a thin unlabeled strip between the two classes. Since these narrow gaps do not correspond to physical land, a simple post-processing step preserves them as background rather than incorrectly labeling them as land. %

Labeling the background as land was not necessary for SAM 3 as it produced semantically coherent segmentations across a wide range of coastal scenes and represents one of the strongest end-to-end foundation-model approaches evaluated. However, the resulting shoreline boundaries were often less precise than those produced by Grounded-SAM~2 or Grounded-MobileSAM, leading to increased boundary localization error despite competitive mean IoU scores. SAM 3 is also among the largest and most computationally expensive models evaluated. Although limited right now, the rapid pace of foundation-model development suggests that future SAM~3-like architectures may eventually eliminate the need for separate semantic and segmentation backends.

Across all evaluated models, a clear trade-off emerges between segmentation accuracy and computational efficiency.
YOLO-based approaches provide the highest frame rates, while transformer-based foundation models generally produce more reliable shoreline boundaries. 
Grounded-MobileSAM achieves the highest overall segmentation accuracy, obtaining the highest mean IoU and lowest shoreline segmentation error. Its reliance on Grounding DINO prompts for every frame limits processing speed because MobileSAM lacks temporal memory between observations. Grounded-SAM~2 achieves nearly identical segmentation accuracy while operating substantially faster by exploiting SAM~2's temporal memory mechanism, allowing Grounding DINO to be invoked only intermittently. 
The 95th-percentile shoreline error increases by only 1.3 pixels relative to Grounded-MobileSAM, a modest penalty given the substantial improvement in runtime performance. Taken together, these results identify Grounded-SAM~2 as the most practical operating point among the evaluated approaches. It therefore serves as the perception frontend for the remainder of this work.

\begin{table}[h]
\centering
\footnotesize
\setlength{\tabcolsep}{2.5pt}
\caption{Visual trajectory accuracy comparison. Metrics are reported as translational error $T_{\mathrm{err}}$ [m] and rotational error $R_{\mathrm{err}}$ [deg] (ATE RMSE). Runtime [min] includes the complete image-processing and optimization pipeline. -- indicates failure to process the entire dataset.}
\label{tab:visual_accuracy_gt}
\begin{tabular}{>{\raggedleft\arraybackslash}p{1.8cm} cccccc}
\toprule
 & Baseline & \multicolumn{3}{c}{Proposed Framework} & \multicolumn{2}{c}{External Baselines} \\
\cmidrule(lr){2-2} \cmidrule(lr){3-5} \cmidrule(lr){6-7}
Metric & \begin{tabular}{@{}c@{}}Odom \\ Only\end{tabular} & \begin{tabular}{@{}c@{}}Fixed \\ Lag\end{tabular} & Hierarchy & \begin{tabular}{@{}c@{}}Full \\ History\end{tabular} & \begin{tabular}{@{}c@{}}VINS \\ Mono\end{tabular} & \begin{tabular}{@{}c@{}}ORB \\ SLAM3\end{tabular} \\
\midrule
\multicolumn{7} {l}{\textbf{Makali\textquoteleft{}i Point} \textit{(18.3 min)}} \\
$T_{\text{err}}$ & 42.9 & 13.9 & 12.1 & 12.3 & -- & -- \\
$R_{\text{err}}$  & 1.4  & 1.4  & 1.7  & 1.4  & -- & -- \\
Runtime & 6.5  & 15.2 & 15.7 & 71.8 & -- & -- \\
\cmidrule(lr){1-7}
\multicolumn{7} {l}{\textbf{K\={a}ne\textquoteleft{}ohe Bay} \textit{(15.0 min)}} \\
$T_{\text{err}}$   & 32.4 & 19.5 & 13.5 & 12.5 & 827 & -- \\
$R_{\text{err}}$  & 1.4  & 1.4  & 1.4  & 1.4  & 13.0 & -- \\
Runtime & 5.6  & 12.4 & 12.7 & 41.0 & 3.8 & -- \\
\cmidrule(lr){1-7}
\multicolumn{7} {l}{\textbf{\textquoteleft{}Anahulu River} \textit{(16.0 min)}} \\
$T_{\text{err}}$   & 41.3 & 16.2 & 11.2 & 9.8  & 327 & 33.2 \\
$R_{\text{err}}$  & 0.4  & 0.5  & 0.5  & 0.6  & 36.1 & 6.3 \\
Runtime & 5.4  & 13.5 & 13.8 & 53.3 & 5.0 & 10.7 \\
\bottomrule
\end{tabular}
\end{table}

\begin{figure}[h]
    \centering
    \includegraphics[width=\columnwidth]{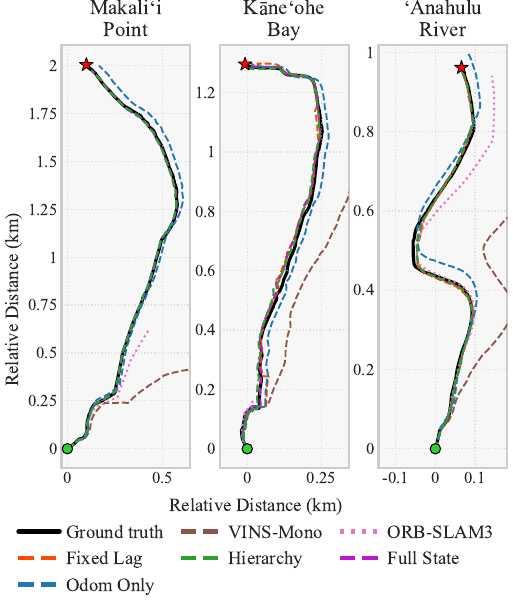}
    \caption{
    Comparison of trajectories produced by the proposed vision pipeline and visual odometry baselines across the three evaluation datasets.
    }
    \label{fig:trajectory_comp_visual}
\end{figure}

\subsection{Vision-Based Localization Results}

With the perception front-end validated, we now evaluate the complete monocular localization framework. Specifically, we seek to quantify both the benefit of shoreline-based satellite alignment and the impact of the proposed hierarchical optimization architecture.

As a baseline, we first consider the local fixed-lag smoother without shoreline registration updates. This estimator is the local odometry layer used for submap construction in Fig.~\ref{fig:visual_fg} and is initialized from the ground-truth pose to isolate drift accumulated during operation.
The baseline is compared against increasingly capable formulations: (1) the local fixed-lag smoother augmented with shoreline-based satellite alignment updates, (2) the proposed hierarchical factor graph architecture, and (3) a full-history optimization. Additionally, VINS-Mono~\cite{vins_mono} and ORB-SLAM3~\cite{ORBSLAM3_TRO} are included as visual-inertial odometry and SLAM benchmarks. The resulting trajectories are shown in Fig.~\ref{fig:trajectory_comp_visual}, with accuracy and runtime metrics summarized in Table~\ref{tab:visual_accuracy_gt}.

Like the LiDAR baselines, the external visual baselines are sensitive to the availability of nearby land features. Neither VINS-Mono nor ORB-SLAM3 successfully processes the Makali\textquoteleft{}i Point dataset. ORB-SLAM3 also fails to complete the K\={a}ne\textquoteleft{}ohe Bay trajectory, while VINS-Mono accumulates substantial drift, resulting in a translational error of 827~m. Performance improves in the more constrained \textquoteleft{}Anahulu River environment, where nearby land provides abundant visual features, yet VINS-Mono and ORB-SLAM3 still produce translational errors of 327~m and 33.2~m. Inspection of the tracked features further reveals that, although most features are detected on land, both methods frequently detect and attempt to track features associated with the water surface. These results reinforce the difficulty of applying conventional visual-inertial methods in coastal environments and highlight the need to exploit unique environmental cues.

Across all three datasets, shoreline submap alignment substantially reduces translational error relative to the local odometry baseline, demonstrating that coastal observations can provide effective global position corrections. Extending the local estimator with the hierarchical global graph yields a further reduction in translational error, particularly in the K\={a}ne\textquoteleft{}ohe Bay and \textquoteleft{}Anahulu River datasets. This behavior directly supports the hallway-problem motivation underlying the proposed hierarchical optimization framework. When informative shoreline geometry is encountered after an extended locally linear segment, the global graph can propagate delayed corrections beyond the limited history retained by the fixed-lag smoother.

In the K\={a}ne\textquoteleft{}ohe Bay and \textquoteleft{}Anahulu River datasets, the monolithic full-history optimization achieves slightly lower translation error than the hierarchical approach, while both methods produce nearly identical results in the Makali\textquoteleft{}i Point dataset. These accuracy gains from maintaining the full trajectory history are modest relative to the additional computational burden. The hierarchical framework, including the complete image-processing pipeline used to generate shoreline submaps, processes each trajectory in approximately 13 to 16 minutes, only about 30 seconds slower than the fixed-lag smoother. 
By comparison, the full-history optimization requires 41 to 72 minutes, reflecting the cost of repeatedly re-optimizing the full trajectory history. 
This tradeoff makes the proposed architecture an attractive middle ground, capturing most of the accuracy benefit of full-history optimization while operating at nearly the same computational cost as the fixed-lag smoother.

\subsection{Ablation Study: The Impact of Water Returns on LiDAR Odometry}

A common explanation for the failure of LiDAR odometry pipelines in maritime environments is the presence of spurious returns generated by wave crests, vehicle wake, and water-surface reflections. To evaluate this hypothesis, we conducted an ablation study in which all LiDAR returns not associated with land were removed prior to odometry estimation. Using the semantic segmentation pipeline and calibrated camera--LiDAR extrinsics, LiDAR points were projected into the image plane and retained only if they intersected the land mask. The resulting land-only point clouds were then used to re-evaluate each LiDAR odometry pipeline in Evalio, with results summarized in Table~\ref{tab:trajectory_comparison}.

\begin{table}[t]
\centering
\footnotesize %
\setlength{\tabcolsep}{3.0pt} %
\caption{LiDAR odometry ablation with semantic land masking constraints. Metrics are presented as translation [m] / rotation [deg] (ATE RMSE). Bold entries denote an improvement over the unmasked baseline (Table~\ref{tab:lidar_trajectory_comparison}), and $\dagger$ marks a regression to failure induced by the land mask.}
\label{tab:trajectory_comparison}
\begin{tabular}{l cc cc cc cc cc}
\toprule
\multirow{2}{*}{Dataset} & \multicolumn{2}{c}{KISS-ICP} & \multicolumn{2}{c}{FORM} & \multicolumn{2}{c}{LOAM} & \multicolumn{2}{c}{LIO-SAM} & \multicolumn{2}{c}{GenZ-ICP} \\
\cmidrule(lr){2-3} \cmidrule(lr){4-5} \cmidrule(lr){6-7} \cmidrule(lr){8-9} \cmidrule(lr){10-11}
 & $T$ & $R$ & $T$ & $R$ & $T$ & $R$ & $T$ & $R$ & $T$ & $R$ \\
\midrule
Makali\textquoteleft{}i Point & 70.0 & 5.0 & \textbf{71} & 13 & 390 & 75 & $\dagger$ & $\dagger$ & \textbf{372} & \textbf{46} \\
K\={a}ne\textquoteleft{}ohe Bay   & \textbf{15.3} & 3.3 & $\dagger$ & $\dagger$ & \textbf{839} & \textbf{119} & $\dagger$ & $\dagger$ & 460 & 41 \\
\textquoteleft{}Anahulu River & \textbf{2.8}           & \textbf{0.5}          & 6.4 & 1.4 & 45.6  & 7.9  & $\dagger$  & $\dagger$ & \textbf{2.6} & \textbf{0.4} \\
\bottomrule
\end{tabular}
\end{table}

The results reveal a clear distinction between the river and open-water coastal environments. In the \textquoteleft{}Anahulu River dataset, restricting the point cloud to land observations produces some of the most accurate trajectory estimates across all evaluated configurations, suggesting that removing water and water-surface reflection returns can improve odometry performance in calm-water conditions when abundant land returns remain available. 

In contrast, the open-water coastal datasets exhibit little benefit from restricting the point cloud to land observations alone. While KISS-ICP achieves a modest reduction in translational error in K\={a}ne\textquoteleft{}ohe Bay, most methods experience little change or continue to fail entirely. 
These results suggest that modern LiDAR odometry pipelines are already reasonably effective at rejecting many spurious water-surface observations through existing correspondence filtering and outlier rejection mechanisms. Instead, the primary challenge appears to be the limited availability of informative shoreline structure. In coastal environments, shoreline returns are often sparse, distant, and visible from only one side of the vessel. Restricting the point cloud to observations that fall within the field of view of the shoreline-facing cameras further reduces the number of available constraints and discards potentially informative land returns outside the camera coverage, including observations behind the vehicle. Consequently, any benefit gained from suppressing water-surface artifacts is often outweighed by the simultaneous loss of valid measurements needed for reliable scan registration.

Taken together, these results suggest that water-surface reflections are a secondary challenge in coastal LiDAR odometry. The more fundamental limitation is the scarcity of dense, informative observations for scan registration.

\begin{figure}[t]
    \centering
    \includegraphics[width=0.95\columnwidth]{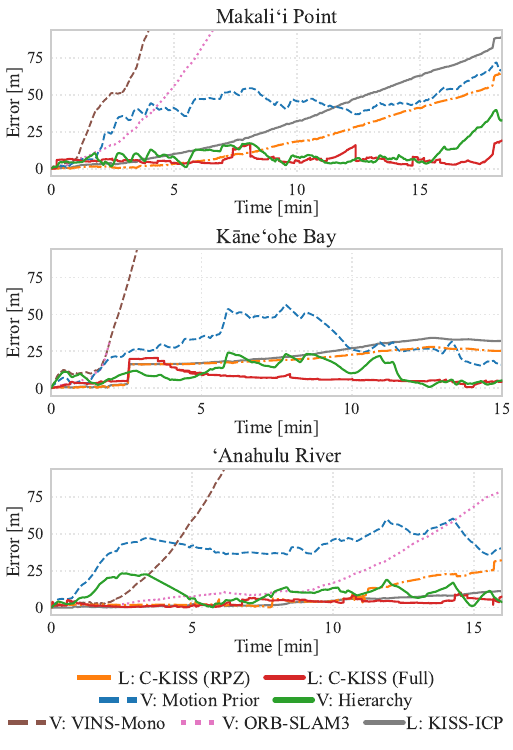}
    \caption{
    Absolute translational trajectory error as a function of mission time for all three datasets. Both the proposed LiDAR and monocular localization frameworks substantially outperform conventional LiDAR and visual-inertial methods. Although trajectory error may temporarily increase during extended sections of locally linear coastline, shoreline-to-satellite registration provides intermittent global corrections that prevent unbounded drift. As a result, both proposed approaches maintain bounded localization error throughout each mission.
    } 
    \label{fig:lidar_error}
\end{figure}

\section{DISCUSSION AND LESSONS LEARNED}
\label{sec:lessons_learned}

The experimental evaluations demonstrate that exploiting coastal scene geometry enables globally consistent localization in GPS-denied maritime environments. However, deploying these pipelines across diverse field settings revealed key insights regarding sensing mechanics, environmental failure modes, and architectural trade-offs between modalities.

\subsection{Maritime Environments Are Not Homogeneous}

Although all three datasets were collected within the same geographic region, the resulting sensing conditions and failure modes differed substantially. Calm inland waterways, sheltered bays, and exposed open coastlines introduce fundamentally different perception challenges because the availability and quality of geometric observations vary significantly between environments. This distinction was most apparent in the calm-water river dataset, where specular reflections from the water surface deceived both estimators.
For the LiDAR pipeline, specular reflections produced returns beneath the apparent water surface, violating the assumed water-plane geometry and corrupting roll, pitch, and height estimates.
For the visual pipeline, mirror reflections led to false shoreline boundaries, causing the estimated land mass to appear artificially closer than its true location.

The river environment remained highly favorable for conventional terrestrial LiDAR odometry, as dense bank vegetation provided abundant features on both sides of the vessel. This contrasts sharply with the open-water coastal datasets, where land returns occurred on only one side of the vessel and at further distances. 
Under these open-water conditions, traditional LiDAR odometry methods routinely failed, reinforcing the necessity of exploiting domain-specific global constraints.
Beyond motivating the need for shoreline-based global constraints, these results demonstrate that observability in maritime environments is highly environment dependent. Localization strategies effective in one maritime setting may not transfer directly to another.

\subsection{Complementary Strengths of LiDAR and Vision}
A primary distinction between the two approaches lies in how they sense and leverage coastal geometry. As an active sensing modality, LiDAR directly measures range, providing highly accurate 3D spatial constraints whenever sufficient shoreline structure is present. These measurements support reliable ego-motion estimation and allow individual shoreline observations to be registered directly against satellite-derived coastline maps. However, unlike the visual pipeline, LiDAR does not readily exploit the horizon as a source of attitude information. Estimating roll and pitch instead requires extracting the coastal boundary and fitting a water-surface plane from the LiDAR observations, making attitude estimation dependent on the quality and visibility of both the water surface and shoreline.

As a passive sensing modality, the visual pipeline obtains roll and pitch directly from observations of the horizon and cannot directly measure range. Instead, shoreline distance is estimated through a geometric reconstruction process that combines shoreline segmentation, camera calibration, and estimated vehicle attitude. Consequently, localization accuracy becomes sensitive to both segmentation quality and attitude estimation errors. Spatial discretization from camera resolution, paired with inherent visual ambiguity along natural shorelines, allows small pixel-level segmentation errors to translate into multi-meter position errors. Furthermore, distant shoreline features occupy fewer pixels, reducing image sharpness and increasing uncertainty in the reconstructed shoreline.

These effects highlight a complementary trade-off between the two sensing modalities. While the visual pipeline yields highly informative attitude constraints from the horizon, LiDAR provides superior spatial fidelity for shoreline reconstruction, ego motion estimation, and coastline registration through direct range sensing. This distinction helps explain why the LiDAR framework consistently achieved lower trajectory error, as the visual framework remained more dependent on accumulated observations and delayed global corrections.

\subsection{Observability Degradation during Prolonged Straightaways}
Beyond the sensing trade-offs discussed above, localized unobservability exposes a critical bottleneck in the visual framework's local motion estimation. 
As shown in Fig.~\ref{fig:lidar_error}, localization error for the visual pipeline grows most rapidly during extended passes of locally linear coastlines. In these regions, shoreline registration provides little information about motion parallel to the coast, allowing position error to accumulate in that direction.

Although the proposed hierarchical factor graph retroactively distributes corrections once a geometrically informative shoreline segment is encountered, error accumulation remains pronounced during prolonged straightaways where the underlying dead-reckoning backbone is weak. This behavior is directly visible in several key segments of Fig.~\ref{fig:lidar_error}, including the final portion of the Makali\'i Point trajectory (15--18~min), the central section of K\={a}ne\'ohe Bay (5--12~min), and the opening segment of the \textquoteleft{}Anahulu River dataset (0--5~min). During development, estimating visual ego-motion directly from shoreline observations proved too noisy to serve as a reliable local odometry source, leaving the system reliant on a weak velocity prior, magnetometer heading, and horizon attitude factors. Incorporating dedicated speed-over-ground sensors, such as a Doppler Velocity Log (DVL), could provide substantially stronger local motion constraints during extended periods of weak shoreline observability.

\subsection{Limitations of Segmentation and Map-Based Registration}
Finally, the visual framework highlights specific vulnerabilities introduced by frontend segmentation assumptions and reference map dependencies. To overcome the tendency of off-the-shelf foundation models to leave portions of the land unlabeled, our vision frontend assumes that all pixels not categorized as water or sky represent land. While effective across the evaluated datasets, this assumption breaks down in the presence of vessels, dynamic or stationary. Furthermore, the cross-view registration relies on structural consistency between live observations and satellite imagery. Significant tidal fluctuations may alter the visible shoreline relative to the satellite map, degrading alignment accuracy and placing an additional burden on the backend smoother.

Across both localization frameworks, the lessons learned suggest that successful maritime localization depends not only on sensing capability, but also on understanding the unique observability characteristics and environmental assumptions of coastal operation.

\section{CONCLUSION}
\label{sec:conclusion}
Rather than seeking general-purpose solutions that merely operate in coastal environments, this paper demonstrates how the shoreline and water surface can be exploited as geometric constraints for globally referenced localization in GPS-denied maritime environments. The proposed LiDAR framework, Coastal-KISS, exploits water-surface observations to estimate roll, pitch, and heave while registering shoreline observations against satellite-derived coastline maps to obtain position and heading. 
The monocular framework instead reconstructs shoreline observations from segmented imagery and accumulates them into local submaps that are registered against the same satellite map to constrain position. 
To address the limited observability of these visual shoreline measurements, the monocular framework employs a hierarchical factor graph architecture that enables delayed corrections to be propagated beyond the horizon of a local fixed-lag smoother.

The proposed methods were evaluated across three real-world coastal datasets spanning nearly 5~km of autonomous vessel operation and encompassing diverse shoreline structure, stand-off distances, and sea states. Experimental results demonstrate that coastal geometry provides a powerful source of information for maritime localization. 
Coastal-KISS consistently improved trajectory accuracy relative to conventional LiDAR odometry pipelines, while the visual framework achieved globally bounded localization through shoreline-to-satellite registration within a hierarchical factor graph.
Additionally, our evaluation of modern foundation-model approaches demonstrated that zero-shot semantic segmentation is sufficiently mature to support coastal localization across diverse maritime environments.

Beyond localization accuracy, this work reveals several important lessons regarding maritime perception. In particular, maritime localization cannot be treated as a single, homogeneous problem. Calm inland waterways, sheltered bays, and exposed coastlines present fundamentally different sensing conditions, each introducing distinct sources of uncertainty and failure. More broadly, these results suggest that maritime environments should not be viewed merely as another deployment setting for terrestrial localization algorithms, but rather as sensing domains with unique geometric structure and observability characteristics.

Future work will investigate tighter integration of the LiDAR and visual pipelines within a unified multi-modal estimator, including the development of LiDAR-based horizon constraints for improved attitude estimation. We also plan to improve robustness to dynamic coastal objects and tidal variation, and to incorporate additional velocity sensing, such as a DVL, to strengthen local motion estimation in the visual framework.

Beyond these specific improvements, our results suggest that, while robotics research increasingly seeks general-purpose solutions, explicitly exploiting domain-specific environmental cues remains a powerful tool for achieving robust and globally consistent localization.

\section*{ACKNOWLEDGMENT}
GTGPT-4 based on ChatGPT 4, ChatGPT 5, and Gemini 3 AI models were used as collaborative tools for code development. All generated code was thoroughly edited and tested by the authors.

\bibliographystyle{IEEEtran}
\bibliography{ref}

%\begin{IEEEbiographynophoto}
%{FIRST A. AUTHOR,} photograph and biography not available at the time
%of publication.
%\end{IEEEbiographynophoto}

%\begin{IEEEbiographynophoto}
%{SECOND B. AUTHOR,} photograph and biography not available at the time
%of publication.
%\end{IEEEbiographynophoto}

\vfill\pagebreak

\end{document}